\documentclass[10pt,twocolumn,letterpaper]{article}

\usepackage{wacv}              %

\usepackage{graphicx}
\usepackage{amsmath}
\usepackage{amssymb}
\usepackage{booktabs}
\usepackage{times}
\usepackage{bbding}
\usepackage{caption}
\usepackage{multirow}
\usepackage[dvipsnames]{xcolor}
\usepackage[accsupp]{axessibility}  %

\usepackage[pdftex,breaklinks,colorlinks, pagebackref, bookmarks=False]{hyperref}

\usepackage[capitalize]{cleveref}
\crefname{section}{Sec.}{Secs.}
\Crefname{section}{Section}{Sections}
\Crefname{table}{Table}{Tables}
\crefname{table}{Tab.}{Tabs.}
\Crefname{equation}{Eq.}{Eqs.}
\newcommand{\sot}{\texttt{[SoT]}\xspace}
\newcommand{\eot}{\texttt{[EoT]}\xspace}

\makeatletter
\renewcommand{\paragraph}{%
  \@startsection{paragraph}{4}%
  {\z@}{0.25em}{-1em}%
  {\normalfont\normalsize\bfseries}%
}
\makeatother

\def\wacvPaperID{1991} %
\def\confName{WACV}
\def\confYear{2024}

\begin{document}

\title{Training-Free Layout Control with Cross-Attention Guidance}

\author{Minghao Chen
\quad
Iro Laina
\quad
Andrea Vedaldi
\\[0.3em]
Visual Geometry Group, University of Oxford\\
{\tt\small \{minghao, iro, vedaldi\}@robots.ox.ac.uk}  \\
\href{https://silent-chen.github.io/layout-guidance/}{\tt\small {\nolinkurl{silent-chen.github.io/layout-guidance}}}
}
\twocolumn[{
\renewcommand\twocolumn[1][]{#1}%
\maketitle
\thispagestyle{empty}
\begin{center}
    \centering
    \includegraphics[width=\linewidth]{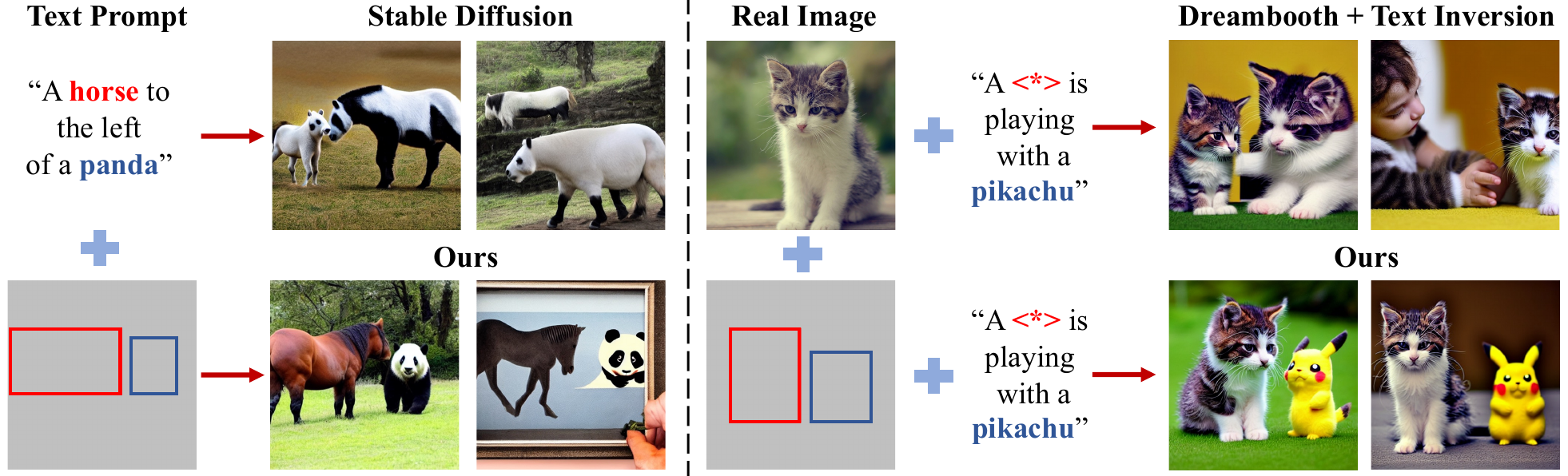}
    \captionsetup{type=figure}
    \captionof{figure}{Left: our method controls the layout of an image generated by a pre-trained diffusion model, such as Stable Diffusion~\cite{stable_diffusion}, without any training or finetuning.
    It also alleviates the problem of such generators omitting certain objects present in the prompt.
    Right: given a single real image, our method can be also used to edit the position and context of a subject (represented by \textcolor{red}{$\langle*\rangle$ }).}
    \label{fig:overview}
\end{center}
}]
\begin{abstract}
Recent diffusion-based generators can produce high-quality images from textual prompts.
However, they often disregard textual %
instructions that specify the spatial layout of the composition.
We propose a simple approach that achieves robust layout control without the need for training or fine-tuning of the image generator. Our technique %
manipulates the cross-attention layers that the model uses to interface textual and visual information and steers the generation in the desired direction given, e.g., a user-specified layout. To determine how to best guide attention, we study the role of attention maps
and explore two alternative strategies, forward and backward guidance.
We thoroughly evaluate our approach on three benchmarks and provide several qualitative examples and a comparative analysis of the two strategies that demonstrate the superiority of backward guidance compared to forward guidance, as well as prior work. 
We further demonstrate the versatility of layout guidance by extending it to applications such as editing the layout and context of real images.
\end{abstract}

\section{Introduction}\label{s:intro}

Generative AI is one of the most disruptive technologies that emerged in the past years.
In computer vision, new text-to-image generation methods, such as DALL-E~\cite{ramesh2021zero}, Imagen~\cite{saharia2022photorealistic}, and Stable Diffusion~\cite{stable_diffusion}, have demonstrated that machines are capable of generating images of a quality high enough for use in numerous applications, multiplying the productivity of professional artists as well as lay people.

Despite this success, however, many practical applications of image generation, particularly in a professional setting, require a high level of \emph{control} that such methods lack.
Specifications in language-based image generators are textual; and while text can tap into a vast library of high-level concepts, it is a poor vehicle for expressing fine-grained visual nuances in an image.
Specifically, text is often inadequate for describing the exact \emph{layout of a composition}.

In fact, as shown in previous work \cite{visor}, %
state-of-the-art image generators struggle to correctly interpret simple layout instructions specified via text.
For example, when prompting such models with a phrase such as \textit{``a dog to the left of a cat"}, the \textit{``left of''} relationship is not always depicted accurately in the generated images. 
In fact, prompts of this nature often cause models to produce erroneous semantics, for example, an image of a cat-dog hybrid.
This limitation is exacerbated by unusual compositions, \eg, \textit{``horse on top of a house''}, which fall outside the typical compositions the model observes during training.

This work provides a better understanding of this limitation and contributes a mechanism to overcome it. 
To this end, we introduce a method that achieves \emph{layout control} without the need for further training of the image generator, while still maintaining the quality of the generated images.

We note that, while layout cannot be easily controlled via textual prompting, one can \emph{intervene} directly in the cross-attention layers, steering the generation in a direction of choice with user-specified inputs, such as bounding boxes, which we refer to as \emph{layout guidance}.
We consider and compare two alternative strategies for such an intervention: ``forward guidance'' and ``backward guidance''.
Forward guidance directly biases the cross-attention layers to shift activations in the desired pattern, letting the model incorporate the guidance via the iterated application of its denoising steps.
Our main contribution is backward guidance, which uses backpropagation to update the image latents to match the desired layout via energy minimization.

While layout control has already received some attention, with some methods following the forward paradigm~\cite{balaji2022ediffi,singh2023high}, we show that backward guidance is a more effective mechanism. 
Our second contribution is then an in-depth investigation of the factors that influence the layout during the image generation process, shedding light on the shortcomings of forward guidance and discussing how backward guidance addresses these.
We show that, while there is an intuitive correlation between different concepts and their visual extent, this correlation is more nuanced than one might think, and, perhaps counter-intuitively, even the special tokens in the prompt (start tokens and padding tokens) contribute to shaping the layout.

Finally, we show that our backward guidance outperforms existing methods and seamlessly integrates into applications such as real-image layout editing.

\section{Related Work}\label{s:related}

\paragraph{Text-to-Image Generation.}

For several years, generative adversarial networks (GANs)~\cite{goodfellow14generative} have been the dominant approach in image generation from textual prompts~\cite{reed2016generative,zhang2017stackgan,zhang2018stackgan++,tao2022df,xu2018attngan,zhang2021cross}. 
Alternative representations for text, such as scene graphs, have also been considered~\cite{johnson2018image}.
More recently, the focus has shifted onto text-conditional autoregressive~\cite{ramesh2021zero,yu2022scaling,gafni2022make,ding2021cogview} and diffusion models~\cite{nichol2021glide,saharia2022photorealistic,stable_diffusion,Dalle-v2,gu2022vector}, with impressive results in generating images of remarkable fidelity, while avoiding common GAN pitfalls such as training instability and mode collapse~\cite{dhariwal2021diffusion}. 
A substantial increase in both the data scale~\cite{Laion-5b} and the size and capabilities of transformer models~\cite{radford2021learning} has played a crucial role in enabling this shift.
Typically, these models are designed to accept a textual prompt as input, which may pose a challenge for accurately conveying all details of the image.
This problem is exacerbated with longer prompts or when describing atypical scenes. 
Recent studies have demonstrated the effectiveness of classifier-free guidance~\cite{ho2022classifier} in improving the faithfulness of the generations with respect to the input prompt.
Others focus on improving compositionality, \eg, by combining multiple diffusion models with different operators~\cite{liu2022compositional}, and attribute binding~\cite{feng2023trainingfree,chefer2023attend}. %

\paragraph{Layout Control in Image Generation.}
Image generation with spatial conditioning is closely related to layout control and typically done with bounding boxes or semantic maps~\cite{sun2019image,sylvain2021object,yang2022modeling,zhao2019image,fan2022frido,park2019semantic}. 
These methods do not use text prompts and rely on a closed-set vocabulary to generate images, \ie, the labels of the training distribution (\eg, COCO~\cite{lin2014microsoft}). 
Recent image-text models such as CLIP~\cite{radford2021learning} are now enabling the extension to open-vocabulary. 
However, the precise layout of a composition is still challenging to convey through text alone; 
even then, the \emph{spatial} fidelity of image generators is extremely limited~\cite{visor}. 
Thus, jointly conditioning on text and layout~\cite{gafni2022make,hinz2019generating,huang2022multimodal} and predicting layout from text~\cite{hong2018inferring} have also been considered. 

Recent works \cite{li2023gligen, yang2022reco, balaji2022ediffi, avrahami2023spatext, bar2023multidiffusion, singh2023high, couairon2023zero, xie2023boxdiff} propose to extend the state-of-the-art Stable Diffusion~\cite{stable_diffusion} with spatial conditioning. 
GLIGEN~\cite{li2023gligen} and ReCo~\cite{yang2022reco} fine-tune the diffusion model with gated self-attention layers and additional regional tokens, respectively. Other works \cite{balaji2022ediffi, bar2023multidiffusion, singh2023high, couairon2023zero, xie2023boxdiff} follow a training-free approach. MultiDiffusion~\cite{bar2023multidiffusion} adopts the idea from~\cite{liu2022compositional} by combining masked noise. eDiff-I~\cite{balaji2022ediffi} and HFG~\cite{singh2023high} share a similar idea with our forward guidance, directly intervening in the cross-attention. 
However, they overlook the significance of special tokens in the process.
Concurrently with our work, ZestGuide~\cite{couairon2023zero} and BoxDiff~\cite{xie2023boxdiff} propose to compute a loss on cross-attention to achieve layout control, which is closer to our backward guidance.
Unlike prior work, we use an objective function that does not rely on precise segmentation masks to be provided by the user, and we provide an in-depth analysis of the factors that affect the layout, and consequently, the behavior of both forward and backward strategies. %
Finally, building on top of diffusion, some recent works show controllable image generation from various other conditioning signals~\cite{bansal2023universal,zhang2023adding,huang2023composer}, such as depth or edge maps.

\paragraph{Diffusion-Based Image Editing.}
Most aforementioned methods lack the ability to control or edit an already generated image, or even the ability to edit real images. 
For example, simply changing a word in the original prompt typically leads to a drastically different generation.
This can be circumvented by providing or generating masks for the objects of interest~\cite{nichol2021glide,couairon2022diffedit}.
Prompt-to-prompt~\cite{hertz2022prompt} addresses this issue with simple text-based edits by exploiting the fact that the cross-attention layers present in most state-of-the-art architectures connect word tokens to the spatial layout of the generated images. 
Text-based image editing can also be achieved through single-image model fine-tuning~\cite{kawar2022imagic,valevski2022unitune}.
However, these approaches, while successful at semantically editing entities can only apply such edits \emph{in-place} and do not allow editing of the spatial layout itself.

\section{Method}\label{s:method}

\newcommand{\x}{\boldsymbol{x}}
\newcommand{\y}{\boldsymbol{y}}
\newcommand{\z}{\boldsymbol{z}}
\newcommand{\bepsilon}{\boldsymbol{\epsilon}}

We consider the problem of \emph{layout-guided} text-to-image generation.
Text-based image generators allow to sample images $\x \in \mathbb{R}^{3\times H \times W}$ from a conditional distribution $p(\x \mid y)$ where $y$ is language description.
Given one such generator off-the-shelf, we wish to steer its output to match a desired layout for the generated composition, \emph{without further training or finetuning}.
In other words, our objective is to investigate whether pre-trained text-to-image generators can adhere to a layout specified by the user during inference, without having been trained with explicit layout conditioning.
In the simplest case, given the text prompt $y$, the index $i$ of a word $y_i$ in the text prompt, and a bounding box $B$, we would like to generate an image $\x$ that contains $y_i$ \emph{inside} $B$, essentially modifying the generator to sample from a new distribution $p(\x \mid y,B,i)$ with additional controls.

\subsection{Preliminaries: Stable Diffusion}%
\label{s:sd}

We first briefly review the technical details of Stable Diffusion (SD)~\cite{stable_diffusion}, a publicly accessible, state-of-the-art text-to-image generator representative of an important class of image generators based on diffusion~\cite{ramesh2021zero,saharia2022photorealistic,stable_diffusion}.
SD consists of an image encoder and decoder, a text encoder, and a denoising network that operates in latent space. 

The text encoder $Y = \phi(y)$ maps the input prompt into a tensor of fixed dimension $Y \in \mathbb{R}^{N \times M}$.
This works by prepending a start symbol \sot to $y$ and appending $N - |y| -1$ padding symbols \eot at the end, to obtain $N$ symbols in total.
Then, the function $\phi$, implemented as a large language model (LLM), takes the padded sequence of words as input and produces a corresponding sequence of token vectors  $Y_i \in \mathbb{R}^M$ with $i \in \{1,\dots, N\}$ as output. %

While not crucial for our discussion, SD's encoding network $h$ maps images $\x$ to corresponding latent codes $\z = h(\x) \in \mathbb{R}^{4\times \frac{H}{s}\times \frac{W}{s}}$, where $s$ divides $H$ and $W$.
The function $h$ is an autoencoder with a left inverse $h^*$, such that $\x = h^*\circ h(\x)$.
The main purpose of this component is to replace the problem of modeling $p(\x \mid y)$ with the problem of modeling $p(\z \mid y)$, reducing the spatial resolution $s$-fold.

A key component of SD is the iterative conditional denoising network $D$.
This network is trained to output a conditional sample $\z \sim p(\z \mid y)$ of the latent code $\z$.
It is designed to take a noised sample
$
\z_t = \alpha_t \z + \sqrt{1-\alpha_t} \bepsilon_t
$,
as input,
where $\bepsilon_t$ is normally distributed noise and $\alpha_t$ is a decreasing sequence, from $\alpha_0\approx 1$ to $\alpha_T\approx 0$, representing the noise schedule.
Then, the network $D$ returns an estimate of the noised sample $\z_t$:
$
D(\z_t,y,t) \approx \bepsilon_t.
$
To sample an image, one first samples $\z_T$, which is normally distributed, and applies $D$ iteratively, to obtain the intermediate codes
$
\z_{T-1},\dots,\z_1,\z_0 \approx \z
$.
Finally, $\z$ is converted back to an image via the image decoder $\x = h^*(\z)$.

There is one final aspect of the SD architecture that is relevant for our work.
While there are several design choices that make the network $D$ work well in practice, the mechanism that is of interest in our investigation is \emph{cross-attention}, %
which connects visual and textual information and allows the generation process to be conditioned on text.
Each cross-attention layer takes an intermediate feature tensor
$
\z^{(\gamma)}\in\mathbb{R}^{C\times \frac{H}{r} \times \frac{W}{r}}
$
as input, where $\gamma$ is the index of the relevant layer in the network, and $r$ is a scaling factor defining the spatial resolution at that level of the representation.
The cross-attention map $A^{(\gamma)}$ associates each spatial location
$
u \in
\{1,\dots,\frac{H}{r}\} \times
\{1,\dots,\frac{W}{r}\}
$
to a token indexed by $i \in \{ 1,\dots, N \}$:
$$
A^{(\gamma)}_{ui}
=
\frac
{
\exp\langle Q_{u}^{(\gamma)}, K_i^{(\gamma)} \rangle
}
{
\sum_{j=1}^N
\exp\langle Q_{u}^{(\gamma)}, K_j^{(\gamma)} \rangle
}, \!
\quad
\boldsymbol{a}_{u}^{(\gamma)}
=
\sum_{i=1}^N A^{(\gamma)}_{ui} V_i^{(\gamma)},
$$
where the value $V_i^{(\gamma)}$ and the key $K_i^{(\gamma)}$ are linear transformations of the token embedding $Y_i$ provided by the textual encoder, $Q^{(\gamma)}$ is a linear transformation of $\z^{(\gamma)}$, and $\boldsymbol{a}_{u}^{(\gamma)}$ is the output of the cross-attention layer.

\begin{figure}[t]
    \centering
    \includegraphics[width=\linewidth]{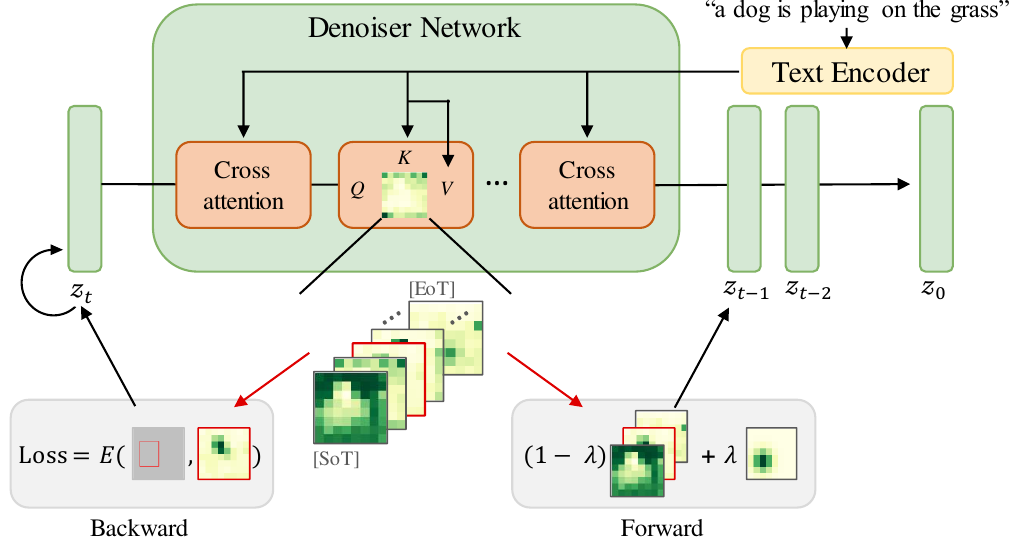}
    \caption{Overview of the two layout guidance strategies. The cross-attention map for a chosen word token is marked with a red border. In forward guidance, the cross-attention maps of the word, start and padding tokens are biased spatially. In backward guidance, we compute instead a loss function and perform backpropagation during the inference process to optimize the latent.}%
    \label{fig:method_overview}
    \vspace{-4mm}
\end{figure}

\begin{figure}[t]
    \centering
    \includegraphics[width=\linewidth]{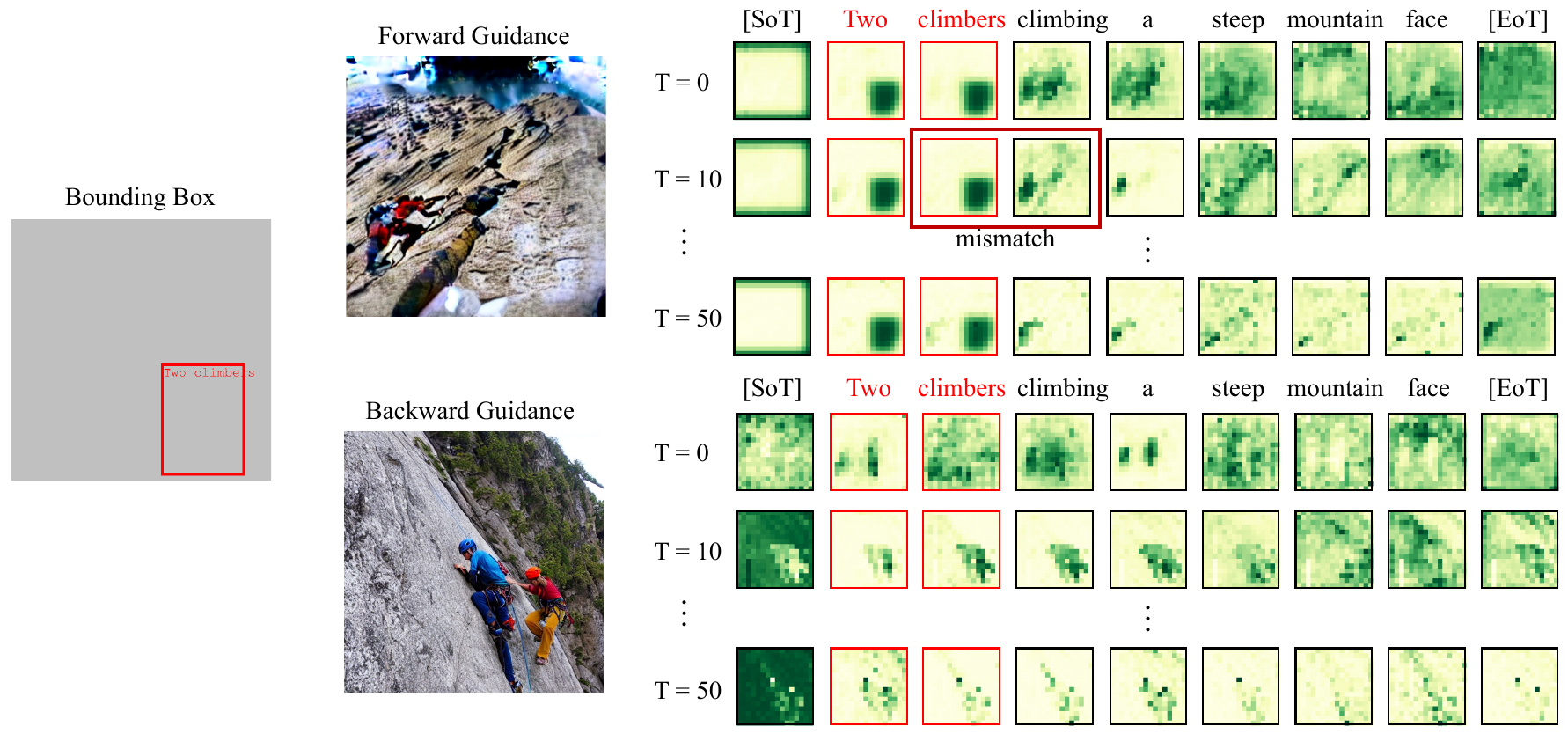}
    \vspace{-1.5em}
    \caption{Cross-attention maps during forward and backward guidance. Spatial dependencies between different words negatively affect forward guidance, while backward guidance softly encourages all dependent tokens to match the desired layout.}
    \label{fig:forward_backward_attention_map}
\vspace{-1mm}
\end{figure}

\subsection{Layout Guidance}%
\label{s:guidance}

Text-to-image generators such as SD struggle to accurately interpret layout instructions provided through text.
We thus introduce a method to guide the layout during the generation process by sampling from a distribution $p(\x \mid y,B,i)$ with additional controls, \eg, user-specified bounding boxes $B$ corresponding to selected text tokens $y_i$. 
This can be achieved via manipulation of the attention response in certain cross-attention layers in the architecture. %

It has already been shown that cross-attention layers regulate the spatial layout of a generated image~\cite{hertz2022prompt}.
Specifically, 
$A^{(\gamma)}_{ui}$ determines how strongly each location $u$ in layer $\gamma$ is associated with each of the $N$ text tokens $y_i$. 
Since the sum of association strengths $\sum_{i=1}^N A^{(\gamma)}_{ui} = 1$ for each spatial location $u$, the different tokens can be seen as ``competing" for a location. %
To control the image layout using a bounding box B corresponding to token $y_i$, the attention can be biased such that locations $u\in B$ within the target box are strongly associated with $y_i$ (while other locations are not). 
As we discuss below, this can be done without fine-tuning the image generator or training additional layers. 

Next, we present a comprehensive investigation of two strategies to achieve training-free layout control: forward and backward guidance (\cref{fig:method_overview}).  
While instances of \textit{forward} guidance have been discussed in recent work~\cite{balaji2022ediffi,singh2023high}, we hereby formalize this approach, identify its limitations, and propose backward guidance as a more effective alternative. 

\paragraph{Forward Guidance.}%
\label{s:forward}

In forward guidance, the bounding box $B$ is represented as a smooth windowing function
$
g^{(\gamma)}_u
$
which is equal to a constant $c > 0$ inside the box and quickly falls to zero outside.%
\footnote{For simplicity, in our implementation, we put a Gaussian blob with $\sigma$ decided by the resolution, height, and width of the bounding box.}
We rescale the windowing function such that $\|g^{(\gamma)}\|_1=1$.
Then, we bias a cross-attention map by replacing it with:
\begin{equation}\label{e:forward}
    A^{(\gamma)}_{ui}
    \leftarrow
    (1-\lambda) A^{(\gamma)}_{ui} +
    \lambda g^{(\gamma)}_u \sum_{v} A^{(\gamma)}_{vi},
    \vspace{-0.5em}
\end{equation}
where $\lambda \in [0,1]$ defines the strength of the intervention.
In practice, we normalize the right side of~\Cref{e:forward} with a softmax function along the text token dimension, keeping the sum of per-pixel attention equal to 1. 
Note that
(1) only the cross-attention map $A^{(\gamma)}_{:,i}$ of the $i$-th token is manipulated,  %
and 
(2) the window is weighed by the mass $\sum_{v} A^{(\gamma)}_{vi}$ %
so as to leave the latter unchanged. 

This intervention is applied for a number of iterations of the denoiser network $D$ at selected layers $\gamma \in \Gamma$.
This means that the activation maps computed by each selected layer are independently modified following~\Cref{e:forward}.  

A critical analysis reveals that forward guidance is a simplistic approach that suffers from inherent constraints hindering its ability to provide effective layout control.
As we discuss in \Cref{s:discussion}, this is primarily due to various factors that influence the layout during the generation process, 
including spatial dependencies among text tokens and spatial information ``hidden'' in the initial noise.

\begin{figure}[t]
    \centering
    \includegraphics[width=0.95\linewidth]{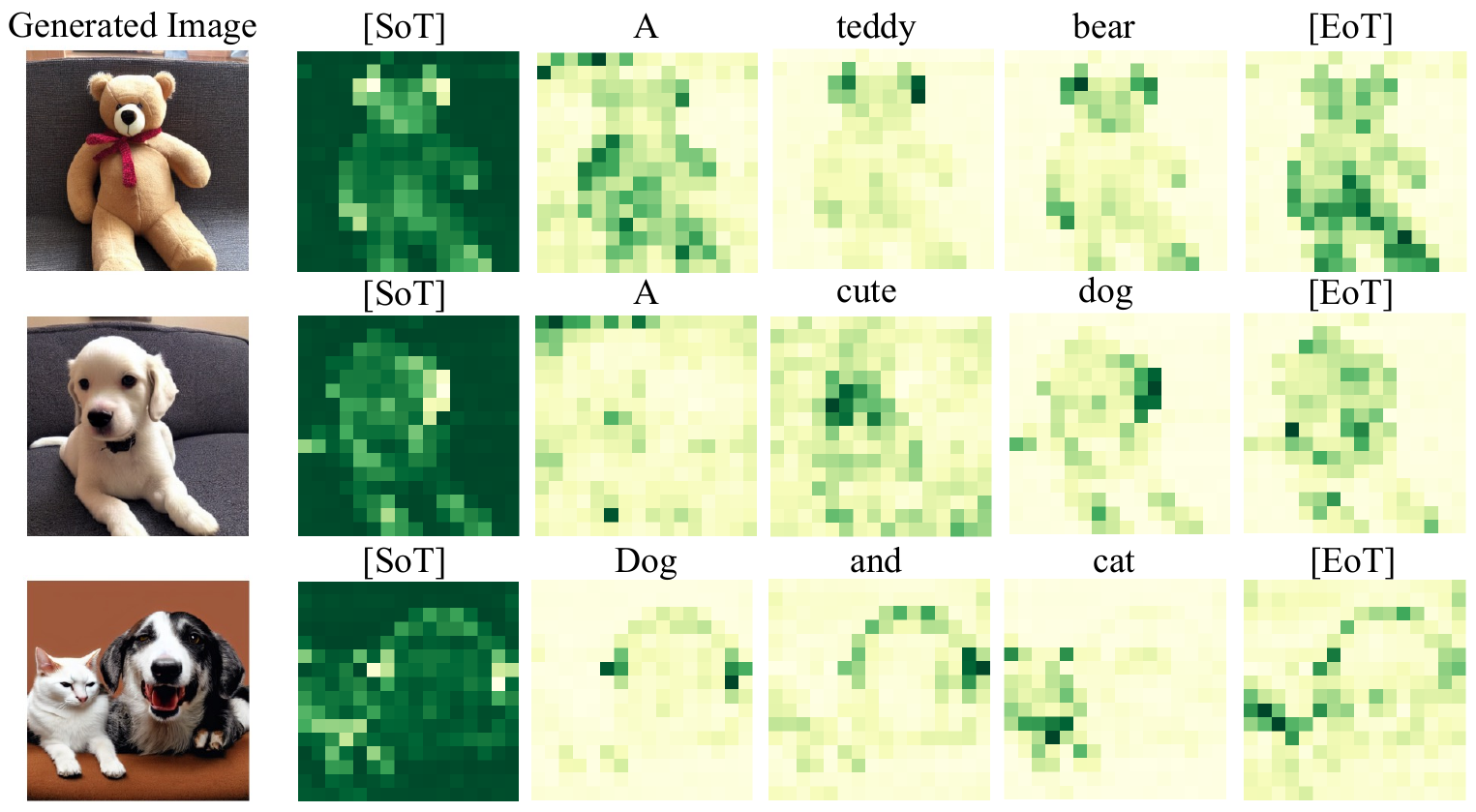}
    \vspace{-2mm}
    \caption{Cross-attention maps of different text prompts at the generation process, indicating that start \sot and padding \eot tokens carry rich semantic and layout information.}
    \label{fig:padding_token}
\vspace{-0.25em}
\end{figure}

\paragraph{Backward Guidance.}%
\label{s:backward}

To address the shortcomings of forward guidance, we introduce an alternative mechanism, which we refer to as backward guidance. %
Instead of directly manipulating attention maps, in backward guidance, we bias the attention by introducing an energy function
\begin{equation}\label{e:backward}
    E(A^{(\gamma)},B,i)
    =
    \left(1 - \frac
    {\sum_{u \in B} A^{(\gamma)}_{ui}}
    {\sum_u A^{(\gamma)}_{ui}}
    \right)^2 .
\end{equation}
Optimizing this function encourages the cross-attention map of the $i$-th token to obtain higher values inside the area specified by $B$.
Specifically, at each application of the denoiser $D$, when layer $\gamma\in\Gamma$ is evaluated, the gradient of the loss~\eqref{e:backward} is computed via backpropagation to update the latent $\z_t (\equiv \z^{(0)}_t)$: %
\begin{equation}
\z_t \leftarrow \z_t - \sigma_t^2 \eta
\nabla_{\z_t} \sum_{\gamma\in\Gamma}E(A^{(\gamma)},B,i) ,
\end{equation}
where $\eta > 0$ is a scale factor controlling the strength of the guidance and $\sigma_t = \sqrt{(1-\alpha_t)/ \alpha_t}$.
By updating the latent, the cross-attention maps of all tokens are indirectly influenced by backward guidance.
To generate an image, we alternate between gradient updates and denoising steps.

\begin{figure}[t]
    \centering
    \includegraphics[width=\linewidth]{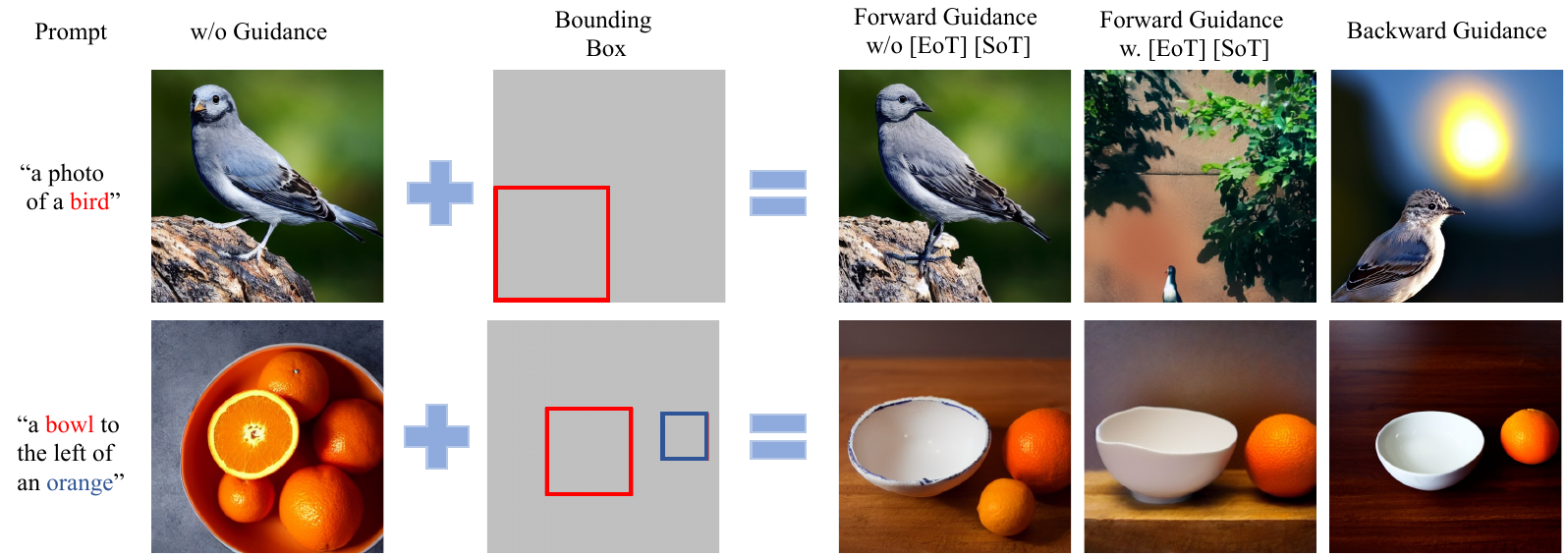}
    \vspace{-5mm}
    \caption{Comparison between forward and backward guidance, including guidance of start and padding tokens.}
    \label{fig:forward_vs_backward}
    \vspace{-5mm}
\end{figure}

\begin{figure*}[t]
    \centering
    \includegraphics[width=\textwidth]{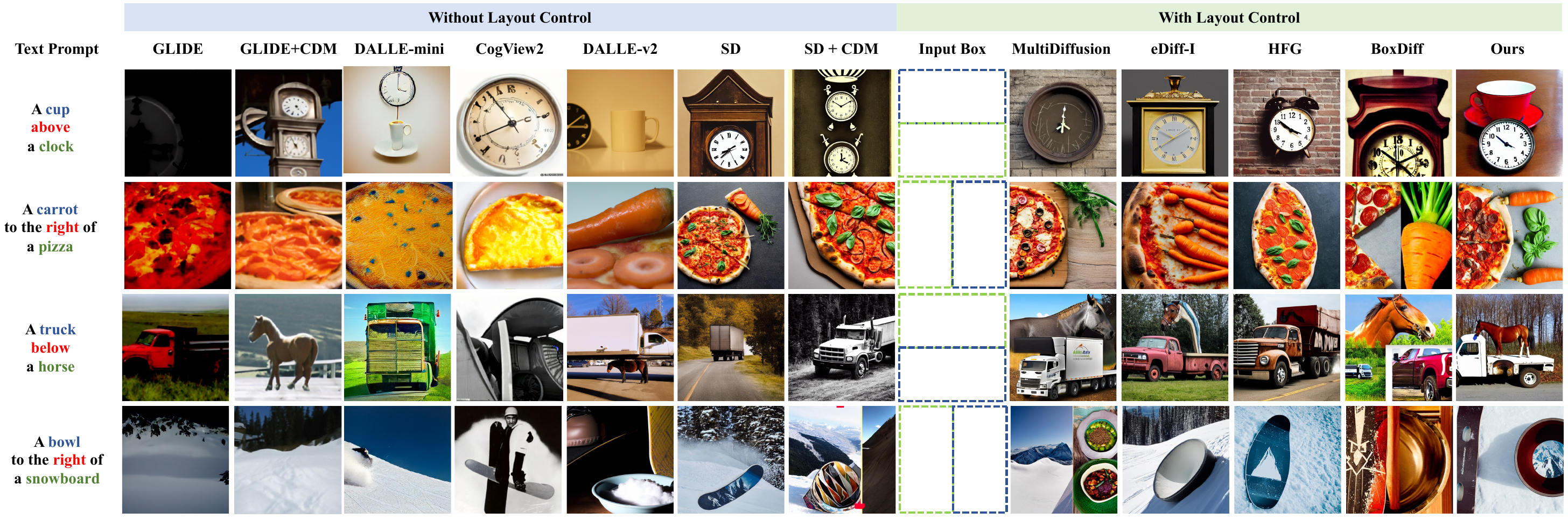}
    \caption{Qualitative comparison of different text-to-image models with text prompts defined in \cite{visor}. As stated in \cite{visor}, current text-to-image models fail to understand spatial relationships without explicit layout conditioning. However, we achieved control of the generated images with the help of guidance on cross-attention maps.}
    \label{fig:Visor_qualitative}
    \vspace{-2mm}
\end{figure*}

\subsection{Analysis and Discussion}%
\label{s:discussion}

Next, we detail a comparative analysis between the forward and backward strategies. 
To motivate backward guidance and understand its effectiveness, we shed light on the significance of all tokens and the influence of the initial noise in shaping the layout during the generation process.

\paragraph{The Role of Word Tokens.}
One important consideration is that the text encoder fuses information from different words when processing a prompt due to self-attention. 
This results in a ``semantic overlap'': information from one token being encoded by another token.
In other words, text embeddings capture both word-specific \emph{and contextual} information, \eg, subject-verb-object dependencies.
This overlap is then transferred from the text encoder into the diffusion process via the cross-attention layers, resulting in \emph{spatial} overlap.
The example in \Cref{fig:forward_backward_attention_map} illustrates this overlap in the cross-attention maps of different words. 
It also shows the behavior of forward and backward guidance when providing spatial conditioning for the phrase ``two climbers''.
It becomes evident that the mismatch between the attention map of the conditioned phrase and its spatial dependencies with other words (``climbing",``a") causes forward guidance to disregard the layout condition.
Instead, backward guidance indirectly drives all attention maps toward the layout condition as necessary, because it acts on the latent codes.

\paragraph{The Role of Special Tokens.}

Another crucial finding is that the cross-attention maps of \sot and \eot tokens, which do not correspond to content words in the input text, still carry significant semantic and layout information.
As we show in \Cref{fig:padding_token}, the cross-attention maps of \eot tokens correspond to salient regions in the generated image, \ie, typically the union of individual semantic entities in the text prompt.
\sot behaves complementarily to \eot, emphasizing the background.
For forward guidance to be effective, it is thus necessary to intervene not only on selected content tokens but also on the special ones.
We use the union of the input boxes as guidance for \eot and the reverse for \sot.
However, we have empirically found that this sometimes results in overly aggressive guidance, which harms image fidelity. 
Backward guidance, on the other hand, does not suffer from such drawbacks, as it optimizes the latent.
We discuss this further in the supplement.

\paragraph{The Role of Initial Noise.}

Finally, the initial noise of the diffusion process plays an important role in shaping the layout of the images. 
We have empirically observed that the noise contains an intrinsic layout; \eg, when prompting the model with phrases like ``an image of a dog'' and ``an image of a cat'' using the same seed, it generates images with consistent layouts, placing the dog and the cat in the same locations. We provide examples in the supplement. 

An initial noise with an intrinsic layout close to the one given by users is easier to optimize and results in higher fidelity. 
Therefore, selecting a noise pattern that aligns with the desired layout can further boost the effectiveness of the guidance.
In backward guidance, the loss applied to the cross-attention maps can, in fact, double as a metric for initial noise selection.
Specifically, we sample different noise patterns and evaluate \Cref{e:backward} after applying backward guidance for a few steps. This allows us to pick the best-aligned initial noise. Please see the supplement for detailed results.

\paragraph{Forward \textit{vs.}~Backward.}

In summary, forward and backward guidance use different mechanisms to manipulate cross-attention. %
Forward guidance \textit{directly} modifies cross-attention to conform to the prescribed pattern, which is ``forced'' repeatedly for a number of denoising iterations. %
While it does not incur any extra computational cost, it struggles to provide robust control over the layout, as non-guided tokens may cause the generation to deviate from the desired pattern.
In contrast, backward guidance uses a loss function to evaluate whether the attention follows the desired pattern. %
While slower than forward guidance, backward guidance is more refined, as it indirectly encourages all tokens (guided and non-guided ones) to adhere to the layout through latent updates. %

\subsection{Real-image Layout Editing}%
\label{s:editing}

Layout guidance can be used in combination with other techniques that build on diffusion-based image generators.
We demonstrate this for the task of real-image editing.
To this end, we incorporate backward guidance into two methods that are commonly used for personalization of diffusion models given real images, namely Textual Inversion (TI)~\cite{text_inversion} and Dreambooth~\cite{dreambooth}.
TI extends an existing image generator with a new concept given one or several images as examples, by optimizing a learnable text token $\langle\ast\rangle$ for the concept.
Dreambooth attempts to capture the appearance of a particular subject of which several images are available by fine-tuning a pre-trained text-to-image model.
Then, new images of the learned concept can be generated.

Neither method supports \textit{localized} spatial control over the newly generated images; their edits are usually global and semantic.
To achieve this, we apply backward guidance on the Dreambooth-finetuned model and the TI-optimized token as part of a prompt. 
This allows us to control the layout of the generated images while preserving the identity of the original object represented by $\langle\ast\rangle$.

\section{Experiments}

In this section, we evaluate our approach for training-free layout guidance, quantitatively comparing variants of forward and backward guidance and providing comparisons to prior and concurrent work on three benchmarks.

\begin{table}[t]
  \centering
  \renewcommand*{\arraystretch}{0.95}
  \footnotesize
    \begin{tabular}{l rr r c}
        \toprule
         & \multirow{2}{*}{\textbf{OA (\%)}} & \multicolumn{2}{c}{\textbf{VISOR (\%)}}  \\
         \cmidrule{3-4}
        \textbf{Model} & & \textbf{uncond} & \textbf{{cond}} & \textbf{Runtime}\\
        \midrule
        Stable Diffusion  &    27.4      &  16.4   &  59.8 & $\sim$ 4 sec/image\\ %
        \midrule

        Ours (FG)     &  25.9      &  23.5   &  90.7 & $\sim$ 4 sec/image\\ %
        Ours (FG$^*$)     &  27.6      &  26.1   &  95.0 & $\sim$ 5 sec/image\\ %
        Ours (BG)    &  38.8     &  37.6     & \textbf{96.9} &  $\sim$ 8 sec/image \\
        Ours (BG + NS) & \textbf{43.7}     &  \textbf{42.3}      & \textbf{96.9} &  $\sim$ 9 sec/image \\
        \bottomrule
    \end{tabular}
    \caption{Comparison of the forward (FG) and backward (BG) strategies, including noise selection (NS). FG$^*$: forward guidance includes \sot and \eot tokens. We randomly sampled 1000 text prompts and compute metrics based on VISOR~\cite{visor}.
    }
    \label{tab:forward_compare_backward}
\end{table}
\begin{table}[t]
  \centering
  \renewcommand*{\arraystretch}{0.95}
  \footnotesize
    \begin{tabular}{l rr r }
        \toprule
         & \multirow{2}{*}{\textbf{OA (\%)}} & \multicolumn{2}{c}{\textbf{VISOR (\%)}} \\
         \cmidrule{3-4}
        \textbf{Model} & & \textbf{uncond} & \textbf{{cond}} \\
        \midrule
        GLIDE \cite{nichol2021glide}            &  3.36      &  1.98      & 59.06 \\ %
        GLIDE + CDM~\cite{liu2022compositional}           &  10.17      &  6.43      & 63.21 \\ %
        DALLE-mini \cite{Dayma_DALLE_Mini_2021}    & 27.10      & 16.17      & 59.67  \\ 
        CogView2 \cite{ding2022cogview2}  & 18.47      & 12.17      & 65.89 \\
        DALLE-v2~\cite{Dalle-v2}     & \textbf{63.93}    & 37.89      & 59.27 \\ %
        SD~\cite{stable_diffusion} & 29.86      & 18.81      & 62.98 \\ %
        SD + CDM~\cite{liu2022compositional}  & 23.27     & 14.99      & 64.41 \\ %
        \midrule
        SD + Ours   & 40.01 & {\bf38.8} & \textbf{95.95}  \\ %
        \bottomrule
    \end{tabular}
    \caption{Comparison of backward guidance (ours) with text-to-image generation models based on the VISOR~\cite{visor} protocol. 
}
    \label{tab:visor_main}
    \vspace{-2mm}
\end{table}
\begin{table} [t!]
  \footnotesize
  \renewcommand*{\arraystretch}{0.95}
  \newcommand{\xpm}[1]{{\tiny$\pm#1$}}
  \centering
\setlength{\tabcolsep}{2.7pt}
\begin{tabular}{@{}lccccc@{}}
  \toprule
  \multirow{2}{*}{Method}  & \multicolumn{2}{c}{COCO 2014} & \multicolumn{3}{c}{Flickr30K} \\
   \cmidrule(lr){2-3} \cmidrule(lr){4-6}
   & FID ($\downarrow$) &mAP ($\uparrow$)  & FID ($\downarrow$) &  AP$_\mathrm{P}$ ($\uparrow$) &mAP ($\uparrow$)  \\
    \midrule
    MultiDiffusion~\cite{bar2023multidiffusion} & 70.7 & 22.3 & 84.1  & 21.6 & 11.9 \\
    eDiff-I~\cite{balaji2022ediffi} & 72.5 & 21.7 & 85.3 & 21.4 & 9.7 \\
    HFG~\cite{singh2023high}& 72.2 & 21.5 & 85.6 & 22.4 & 10.7 \\ 
    BoxDiff~\cite{xie2023boxdiff} & 72.6 & 24.1 & 78.7 & 26.0 & 16.6 \\
    
  \midrule
    Stable Diffusion~\cite{stable_diffusion} & 72.3 & 19.2 & \textbf{76.4} & 19.4 & 8.7  \\
    Stable Diffusion + Ours & 73.3 & 35.7 & 78.9 & 35.6 & 17.9 \\
    \midrule
    GLIGEN~\cite{li2023gligen} & 69.1 & 62.8 & 77.3 & 87.2 & 31.4 \\ 
    GLIGEN + Ours & \textbf{66.7}  & \textbf{65.1} & 78.1 & \textbf{88.9} & \textbf{32.7} \\ 
           
  \bottomrule
\end{tabular}
\caption{Comparison with other layout-to-image models. Our approach improves spatial fidelity (suggested by higher AP/mAP scores). mAP is calculated with an IoU threshold of 0.3.}
\label{table:flickr30k}
\vspace{-3mm}
\end{table}
\subsection{Experimental setup}
\label{s:exp_setup}

\paragraph{Implementation Details.}

We utilize Stable-Diffusion (SD) V-1.5~\cite{stable_diffusion} trained on the LAION-5B dataset \cite{Laion-5b} as the default pre-trained image generator, if not specified.
For a detailed description of the architecture and noise scheduler please see the supplement. %

For forward guidance, we apply \Cref{e:forward} to every layer of the denoiser network for the first 40 steps of the diffusion process and set $\lambda = 0.8$.
For backward guidance, we calculate the loss on the cross-attention maps of the mid-block and the first block of the up-sampling branch of the denoising network (U-Net~\cite{unet}) as we found this to be the optimal setting to balance control and fidelity.
We set $\eta = 30$ by default but found that values between 30-50 work well across most settings.
Since the layout of the generated image is typically established in the early stages of inference, backward guidance is performed during the initial 10 steps of the diffusion process and repeated 5 times at each step.

\paragraph{Evaluation Benchmarks.}

We quantitatively evaluate our approach on three benchmarks: VISOR~\cite{visor}, COCO 2014~\cite{lin2014microsoft}, and Flickr30K Entities~\cite{Flickr30K-entities, Flickr30k}. We discuss the ethical concerns of the dataset usage in the supp. 
VISOR proposes metrics to quantify the spatial understanding abilities of text-to-image models.
For COCO 2014, we follow the same setup adopted by prior work~\cite{bar2023multidiffusion}, which uses only a subset of the annotated objects per image.
Finally, we introduce the Flickr30K Entities dataset as another benchmark to evaluate layout control, since it contains image-caption pairs with visual grounding. 
Details for all benchmarks and metrics are provided in the supplementary material.

\begin{figure}[t]
    \centering
    \includegraphics[width=0.95\linewidth]{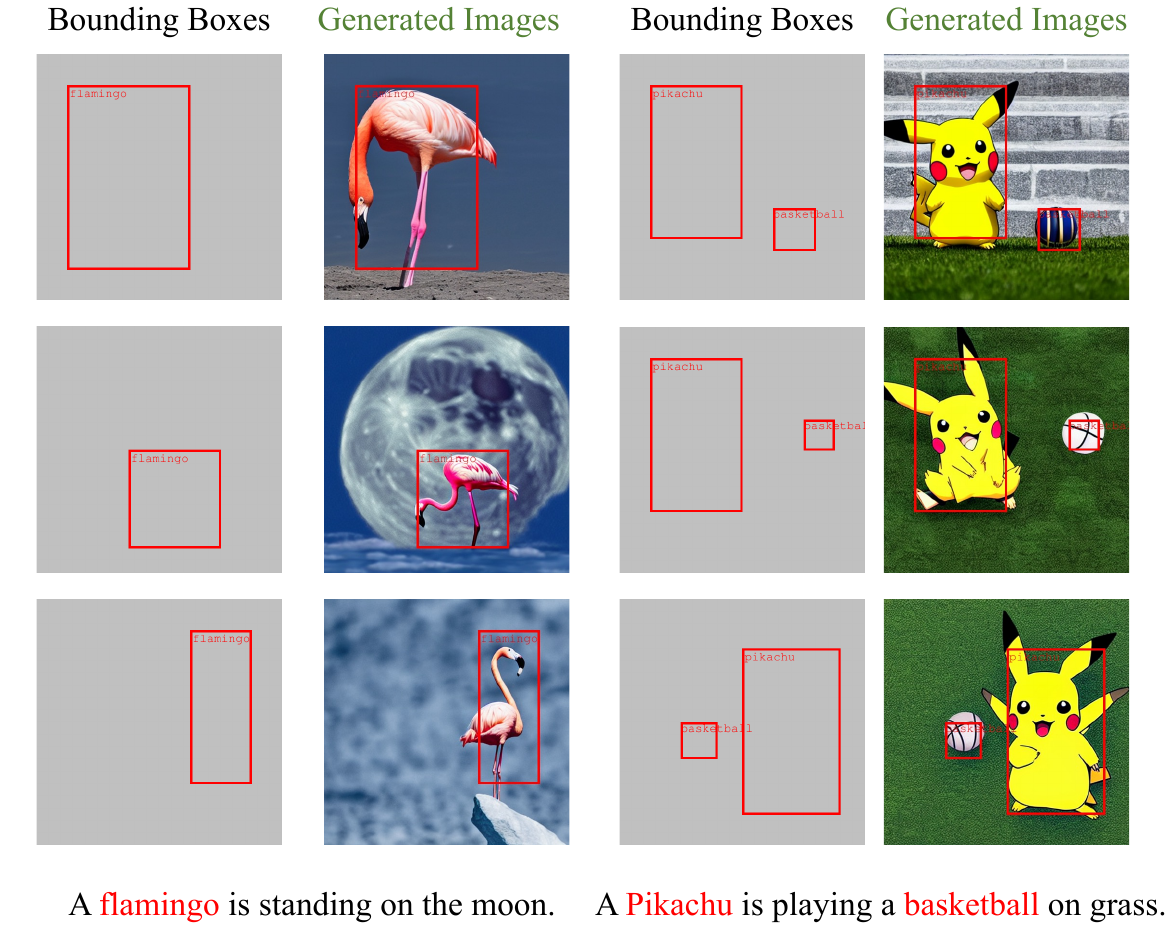}
    \vspace{-0.5em}
    \caption{Our method controls the objects inside the generated images with user-specified bounding boxes. On the left, the size and position of \textit{flamingo} changes according to the bounding box. On the right, we show the ability to control multiple objects.}
    \label{fig:bounding_box}
\end{figure}

\begin{figure}[t]
    \centering
    \includegraphics[width=0.9\linewidth]{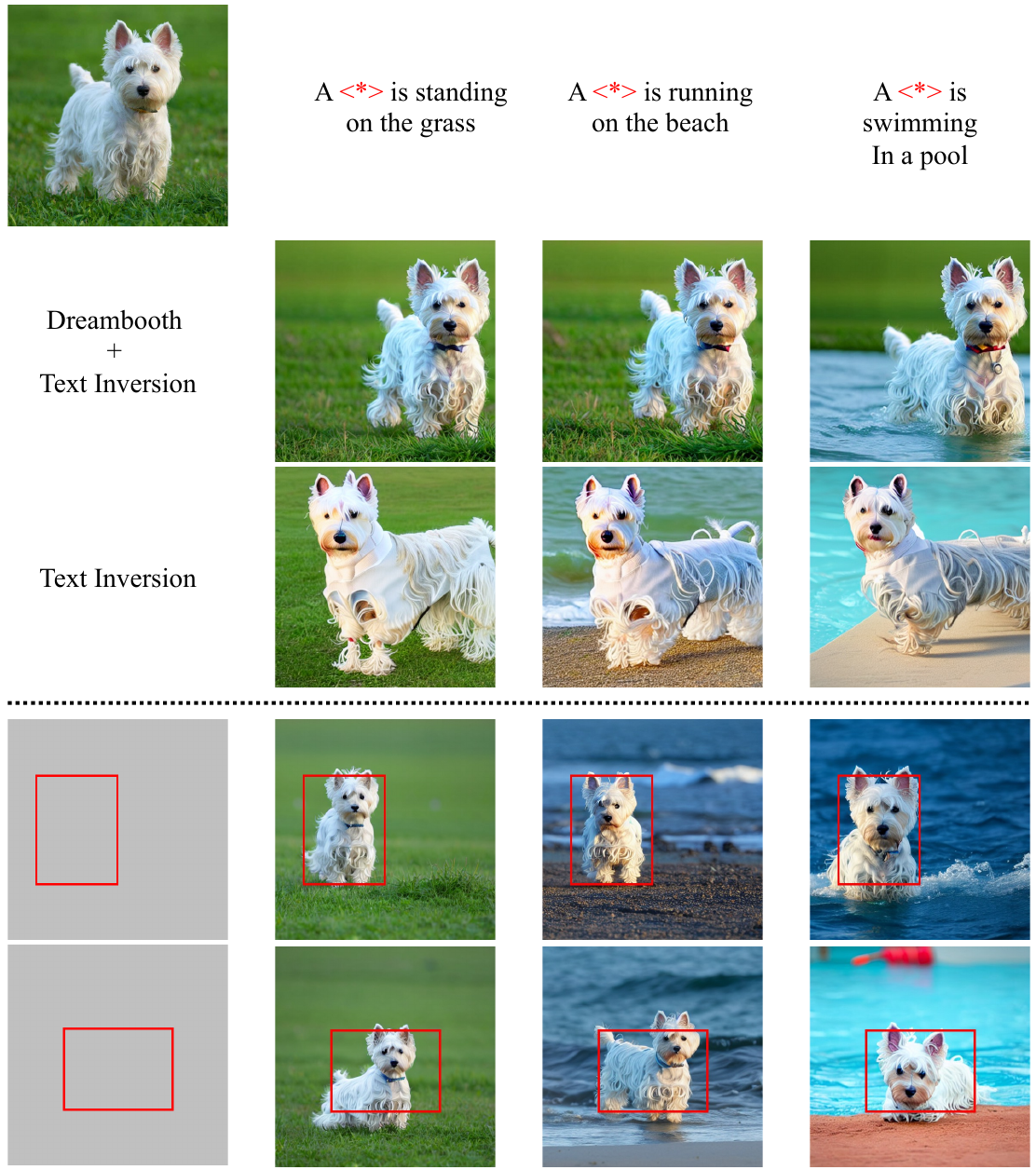}
    \caption{The top left is the real image input. The images above the dash are generations using only text inversion (TI)~\cite{text_inversion} and Dreambooth~\cite{dreambooth}. The images under the line are generated by our method on top of Dreambooth and TI. %
}
    \label{fig:real_image_control}
    \vspace{-4mm}
\end{figure}

\subsection{Forward \textit{vs.}~Backward Guidance}%
\label{s:comparison_guidance}

First, we compare the two different modes of guidance (forward and backward) in \Cref{tab:forward_compare_backward} using the VISOR protocol with 1,000 randomly chosen text samples.
The biggest advantage of forward guidance is that the computation overhead is negligible, thus leading to a faster inference time. %
However, we observe that, compared to (unguided) SD, forward guidance does not significantly increase the object accuracy (OA), while the backward mechanism yields a notably higher OA.  %
In terms of evaluating the generated spatial relationships (VISOR conditional/unconditional metrics), both forward and backward guidance obtain significantly better results than the SD baseline.
We also find that the inclusion of \sot and \eot tokens improves forward guidance, which confirms our analysis and insights in \Cref{s:discussion}, yet backward guidance still achieves superior performance. 
Finally, noise selection using backward guidance offers a significant boost in all metrics.

We provide a qualitative comparison of the forward and backward mechanisms in \Cref{fig:forward_vs_backward}, including the impact of special tokens on forward guidance. 
Backward guidance achieves a better alignment between the generated objects and the input bounding boxes. 
It also helps to address the issue of objects occasionally being omitted from the generated images in diffusion models.

\subsection{Comparisons to Prior Work}%
\label{s:results}

In \Cref{tab:visor_main}, we compare our method with text-to-image generation methods that do not use layout control.
We note that comparisons are fair since, in this setting (VISOR), manual user input is not required for guidance (see supplement). 
Our method exhibits remarkable performance under the VISOR$_\mathrm{cond}$ metric, achieving an accuracy of 95.95\%, and higher OA compared to the baseline (SD).
Although OA does not directly assess layout, the improvement can be explained by the fact that unguided SD often fails to generate correct semantics in atypical compositions. 
We also note that, while DALLE-v2~\cite{Dalle-v2} achieves the highest OA overall, it appears to struggle more with layout instructions compared to SD, as indicated by a lower VISOR$_\mathrm{cond}$ score.

In \Cref{table:flickr30k}, we compare our backward guidance to other mechanisms for layout conditioning. 
Apart from the entries in the last two rows, all methods are based on Stable Diffusion~\cite{rombach2022high} V1.5. Remarkably, our backward guidance surpasses other layout conditioning methods by a significant margin, achieving over a 9-point improvement in mAP and AP$_\mathrm{P}$ on COCO and Flickr30K. Notably, in direct comparison with the concurrent BoxDiff model~\cite{xie2023boxdiff}, we achieve gains of 11.6 in mAP and 9.6 in AP$_\mathrm{P}$, all while maintaining analogous image quality. Finally, we show that our approach can be used complementarily to methods like GLIGEN~\cite{li2023gligen} that train additional layers for layout conditioning, further improving their performance.

In \Cref{fig:Visor_qualitative}, we qualitatively compare different text-to-image models using prompts sampled from \cite{visor}. 
Methods that do not use layout control are not capable of inferring the spatial relationships between objects based purely on textual input and often fail to generate one or both objects.
We also observe that even methods with layout conditioning struggle in this setting, especially those that adopt a forward guidance paradigm (eDiff-I~\cite{balaji2022ediffi}, HFG~\cite{singh2023high}).
In the case of BoxDiff~\cite{xie2023boxdiff}, the lower quality could potentially be due to overlooking the impact of special tokens and the loss function design.
In contrast, our approach (backward-guided SD) can accurately position objects within a scene, even when they are rarely seen together, such as ``snowboard" and ``bowl", and achieves the best adherence to the prompt without loss of image fidelity. 
More examples of our approach are shown in \Cref{fig:bounding_box}, demonstrating precise control over the \emph{size} and \emph{position} of one or more objects, including unconventional object categories, such as ``flamingo'' or ``pikachu'', and atypical scene compositions.

\begin{figure}[t]
    \centering
    \includegraphics[width=\linewidth]{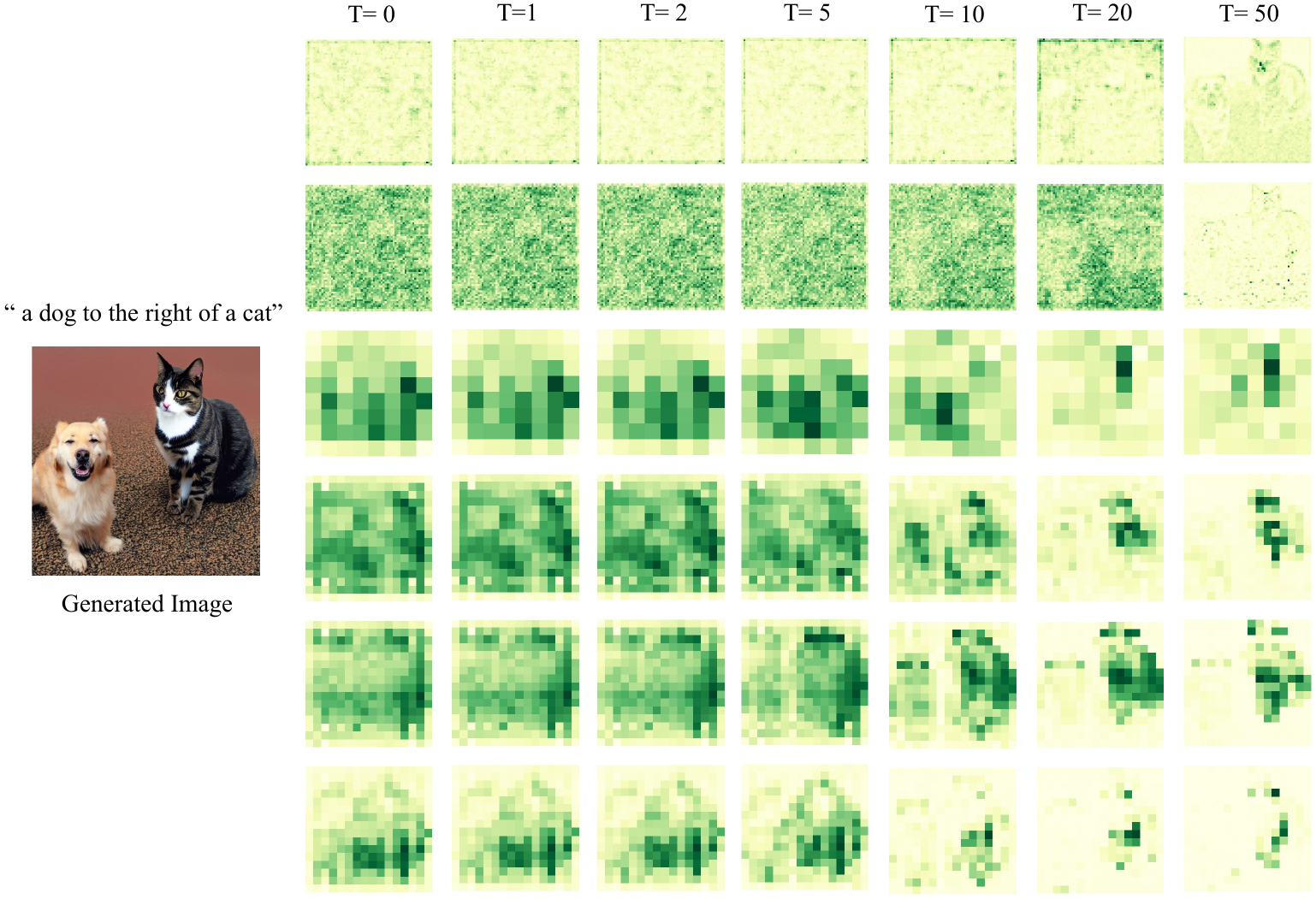}
    \caption{The cross-attention map of the word ``cat" at different layers (top to bottom) across different timesteps (left to right). 
    }
    \label{fig:layer_and_index}
\end{figure}

\begin{figure}[t]
    \centering
    \includegraphics[width=0.99\linewidth]{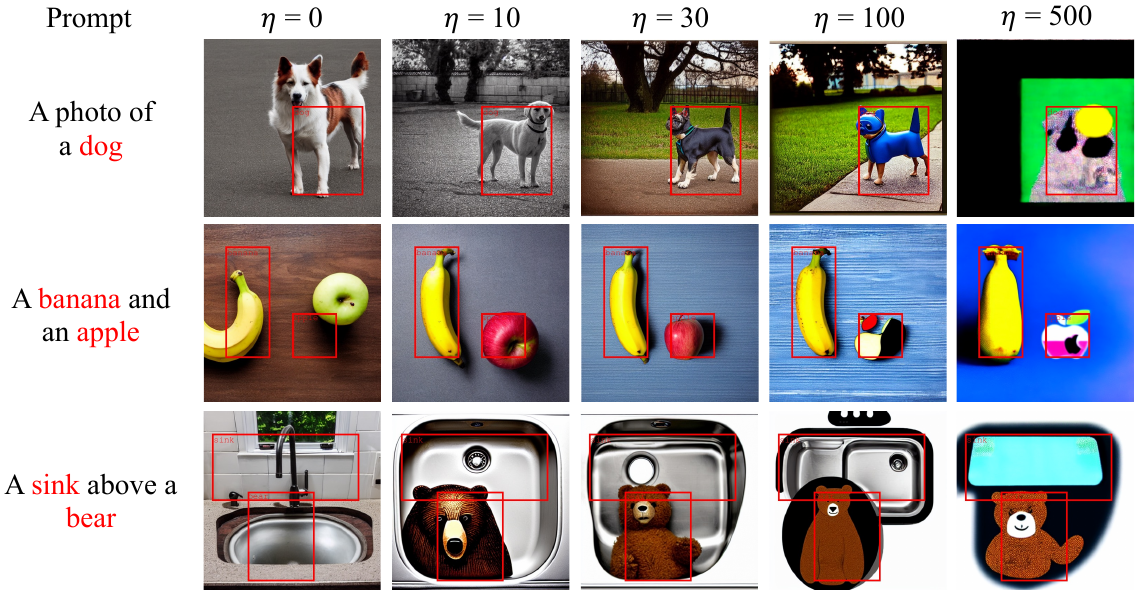}
    \caption{Qualitative comparison of different loss scales in the backward guidance. We increase the loss scale from left to right keeping the same prompt and random seed. With increasing scale, the objects are more tightly constrained inside the bounding boxes. However, for very high scales, fidelity decreases significantly. }
    \label{fig:loss scale}
    \vspace{-2mm}
\end{figure}

\subsection{Further Analysis and Applications}
\label{s:analysis}

\paragraph{Real-Image Layout Editing.}

We showcase the potential of backward layout guidance for editing real images in \Cref{fig:real_image_control}, confirming its effectiveness at changing the position, gesture, and orientation of the ``dog" (based on the aspect ratio of the bounding box) to fit the new context, without altering its identity. 
As shown in the same figure, the capability to precisely control object size and position cannot be attained through Dreambooth/TI alone.
This highlights the potential of our method in a wide range of applications related to image editing and manipulation.

\paragraph{Cross-attention Layers and Guidance Steps.}
We also investigate the layers and the number of guidance steps that are necessary to achieve layout control.
Cross-attention maps at various layers of the denoising network are presented in \Cref{fig:layer_and_index}. 
We observe that the first layers of (down-sampling) do not capture much information about the object (here, the ``cat''). 
We found it most effective to perform backward guidance only on the mid and up-sampling blocks of the architecture. 
The figure also illustrates that object outlines are typically generated in the early steps of the diffusion process, before $T=20$. 
Based on our experimentation, we find that 10-20 steps are generally suitable for guidance. 
Additional quantitative analysis and examples are presented in the supplement.

\paragraph{Loss Scale Factor.}

In \Cref{fig:loss scale}, we qualitatively analyze the impact of the loss scale factor $\eta$. 
We observe that increasing the loss weight leads to stronger control over the generated images, but at the cost of some fidelity, particularly with higher scales.
 The optimal loss scale setting depends on the difficulty of the text prompt. 
 For example, an atypical prompt like \textit{``a sink above a bear"} requires stronger guidance to generate both objects successfully (without guidance, \ie, $\eta = 0$, the bear is not generated). 
 This suggests that layout guidance helps the generator ``recognize'' multiple objects in the text prompt.

\section{Conclusions}\label{s:conclusions}

In this paper, we investigated the potential of manipulating the spatial layout of images generated by large, pre-trained text-to-image models without additional training or fine-tuning. 
Through our exploration, we discovered that both the cross-attention maps and the initial noise of the diffusion play a dominant role in determining the layout and that even the cross-attention maps of special tokens contain valuable semantic and spatial information.
We identify and analyze the mechanism behind most prior work: forward guidance. Moreover, based on our analysis, we propose a new technique ``backward guidance'' that overcomes the shortcomings of forward guidance.
Finally, we demonstrate the versatility of our training-free strategy by extending it to applications such as real-image layout editing.

\paragraph{Ethics.} We use the Flick30K Entities and MS-COCO datasets in a manner compatible with their terms.
Some of these images may accidentally contain faces or other personal information, but we do not make use of these images or image regions.
For further details on ethics, data protection, and copyright please see \url{https://www.robots.ox.ac.uk/~vedaldi/research/union/ethics.html}.

\paragraph{Acknowledgements.}

This research is supported by ERC-CoG UNION 101001212.

{\small

{\small\bibliographystyle{ieee_fullname}\bibliography{arxiv}}

\begin{thebibliography}{10}\itemsep=-1pt

\bibitem{avrahami2023spatext}
Omri Avrahami, Thomas Hayes, Oran Gafni, Sonal Gupta, Yaniv Taigman, Devi Parikh, Dani Lischinski, Ohad Fried, and Xi Yin.
\newblock Spatext: Spatio-textual representation for controllable image generation.
\newblock In {\em Proceedings of the IEEE/CVF Conference on Computer Vision and Pattern Recognition}, pages 18370--18380, 2023.

\bibitem{balaji2022ediffi}
Yogesh Balaji, Seungjun Nah, Xun Huang, Arash Vahdat, Jiaming Song, Karsten Kreis, Miika Aittala, Timo Aila, Samuli Laine, Bryan Catanzaro, et~al.
\newblock ediffi: Text-to-image diffusion models with an ensemble of expert denoisers.
\newblock {\em arXiv preprint arXiv:2211.01324}, 2022.

\bibitem{bansal2023universal}
Arpit Bansal, Hong-Min Chu, Avi Schwarzschild, Soumyadip Sengupta, Micah Goldblum, Jonas Geiping, and Tom Goldstein.
\newblock Universal guidance for diffusion models.
\newblock {\em arXiv preprint arXiv:2302.07121}, 2023.

\bibitem{bar2023multidiffusion}
Omer Bar-Tal, Lior Yariv, Yaron Lipman, and Tali Dekel.
\newblock Multidiffusion: Fusing diffusion paths for controlled image generation.
\newblock {\em arXiv preprint arXiv:2302.08113}, 2023.

\bibitem{chefer2023attend}
Hila Chefer, Yuval Alaluf, Yael Vinker, Lior Wolf, and Daniel Cohen-Or.
\newblock Attend-and-excite: Attention-based semantic guidance for text-to-image diffusion models.
\newblock {\em ACM Transactions on Graphics (TOG)}, 42(4):1--10, 2023.

\bibitem{couairon2023zero}
Guillaume Couairon, Marl{\`e}ne Careil, Matthieu Cord, St{\'e}phane Lathuili{\`e}re, and Jakob Verbeek.
\newblock Zero-shot spatial layout conditioning for text-to-image diffusion models.
\newblock {\em arXiv preprint arXiv:2306.13754}, 2023.

\bibitem{couairon2022diffedit}
Guillaume Couairon, Jakob Verbeek, Holger Schwenk, and Matthieu Cord.
\newblock Diffedit: Diffusion-based semantic image editing with mask guidance.
\newblock {\em arXiv preprint arXiv:2210.11427}, 2022.

\bibitem{Dayma_DALLE_Mini_2021}
Boris Dayma, Suraj Patil, Pedro Cuenca, Khalid Saifullah, Tanishq Abraham, Ph{\'u}c Le~Khac, Luke Melas, and Ritobrata Ghosh.
\newblock Dall{\textperiodcentered} e mini, 2021.

\bibitem{dhariwal2021diffusion}
Prafulla Dhariwal and Alexander Nichol.
\newblock Diffusion models beat gans on image synthesis.
\newblock {\em Advances in Neural Information Processing Systems}, 34:8780--8794, 2021.

\bibitem{ding2021cogview}
Ming Ding, Zhuoyi Yang, Wenyi Hong, Wendi Zheng, Chang Zhou, Da Yin, Junyang Lin, Xu Zou, Zhou Shao, Hongxia Yang, et~al.
\newblock Cogview: Mastering text-to-image generation via transformers.
\newblock {\em Advances in Neural Information Processing Systems}, 34:19822--19835, 2021.

\bibitem{ding2022cogview2}
Ming Ding, Wendi Zheng, Wenyi Hong, and Jie Tang.
\newblock Cogview2: Faster and better text-to-image generation via hierarchical transformers.
\newblock {\em Advances in Neural Information Processing Systems}, pages 16890--16902, 2022.

\bibitem{fan2022frido}
Wan-Cyuan Fan, Yen-Chun Chen, DongDong Chen, Yu Cheng, Lu Yuan, and Yu-Chiang~Frank Wang.
\newblock Frido: Feature pyramid diffusion for complex scene image synthesis.
\newblock {\em arXiv preprint arXiv:2208.13753}, 2022.

\bibitem{feng2023trainingfree}
Weixi Feng, Xuehai He, Tsu-Jui Fu, Varun Jampani, Arjun~Reddy Akula, Pradyumna Narayana, Sugato Basu, Xin~Eric Wang, and William~Yang Wang.
\newblock Training-free structured diffusion guidance for compositional text-to-image synthesis.
\newblock In {\em The Eleventh International Conference on Learning Representations}, 2023.

\bibitem{gafni2022make}
Oran Gafni, Adam Polyak, Oron Ashual, Shelly Sheynin, Devi Parikh, and Yaniv Taigman.
\newblock Make-a-scene: Scene-based text-to-image generation with human priors.
\newblock In {\em Computer Vision--ECCV 2022: 17th European Conference, Tel Aviv, Israel, October 23--27, 2022, Proceedings, Part XV}, pages 89--106. Springer, 2022.

\bibitem{text_inversion}
Rinon Gal, Yuval Alaluf, Yuval Atzmon, Or Patashnik, Amit~H Bermano, Gal Chechik, and Daniel Cohen-Or.
\newblock An image is worth one word: Personalizing text-to-image generation using textual inversion.
\newblock {\em arXiv preprint arXiv:2208.01618}, 2022.

\bibitem{visor}
Tejas Gokhale, Hamid Palangi, Besmira Nushi, Vibhav Vineet, Eric Horvitz, Ece Kamar, Chitta Baral, and Yezhou Yang.
\newblock Benchmarking spatial relationships in text-to-image generation.
\newblock {\em arXiv preprint arXiv:2212.10015}, 2022.

\bibitem{goodfellow14generative}
Ian~J. Goodfellow, Jean Pouget{-}Abadie, Mehdi Mirza, Bing Xu, David Warde{-}Farley, Sherjil Ozair, Aaron~C. Courville, and Yoshua Bengio.
\newblock Generative adversarial nets.
\newblock In {\em Proceedings of Advances in Neural Information Processing Systems (NeurIPS)}, 2014.

\bibitem{gu2022vector}
Shuyang Gu, Dong Chen, Jianmin Bao, Fang Wen, Bo Zhang, Dongdong Chen, Lu Yuan, and Baining Guo.
\newblock Vector quantized diffusion model for text-to-image synthesis.
\newblock In {\em Proceedings of the IEEE/CVF Conference on Computer Vision and Pattern Recognition}, pages 10696--10706, 2022.

\bibitem{hertz2022prompt}
Amir Hertz, Ron Mokady, Jay Tenenbaum, Kfir Aberman, Yael Pritch, and Daniel Cohen-Or.
\newblock Prompt-to-prompt image editing with cross attention control.
\newblock {\em arXiv preprint arXiv:2208.01626}, 2022.

\bibitem{hinz2019generating}
Tobias Hinz, Stefan Heinrich, and Stefan Wermter.
\newblock Generating multiple objects at spatially distinct locations.
\newblock {\em arXiv preprint arXiv:1901.00686}, 2019.

\bibitem{ho2022classifier}
Jonathan Ho and Tim Salimans.
\newblock Classifier-free diffusion guidance.
\newblock {\em arXiv preprint arXiv:2207.12598}, 2022.

\bibitem{hong2018inferring}
Seunghoon Hong, Dingdong Yang, Jongwook Choi, and Honglak Lee.
\newblock Inferring semantic layout for hierarchical text-to-image synthesis.
\newblock In {\em Proceedings of the IEEE conference on computer vision and pattern recognition}, pages 7986--7994, 2018.

\bibitem{huang2023composer}
Lianghua Huang, Di Chen, Yu Liu, Yujun Shen, Deli Zhao, and Jingren Zhou.
\newblock Composer: Creative and controllable image synthesis with composable conditions.
\newblock {\em arXiv preprint arXiv:2302.09778}, 2023.

\bibitem{huang2022multimodal}
Xun Huang, Arun Mallya, Ting-Chun Wang, and Ming-Yu Liu.
\newblock Multimodal conditional image synthesis with product-of-experts gans.
\newblock In {\em Computer Vision--ECCV 2022: 17th European Conference, Tel Aviv, Israel, October 23--27, 2022, Proceedings, Part XVI}, pages 91--109. Springer, 2022.

\bibitem{johnson2018image}
Justin Johnson, Agrim Gupta, and Li Fei-Fei.
\newblock Image generation from scene graphs.
\newblock In {\em Proceedings of the IEEE conference on computer vision and pattern recognition}, pages 1219--1228, 2018.

\bibitem{kawar2022imagic}
Bahjat Kawar, Shiran Zada, Oran Lang, Omer Tov, Huiwen Chang, Tali Dekel, Inbar Mosseri, and Michal Irani.
\newblock Imagic: Text-based real image editing with diffusion models.
\newblock {\em arXiv preprint arXiv:2210.09276}, 2022.

\bibitem{li2023gligen}
Yuheng Li, Haotian Liu, Qingyang Wu, Fangzhou Mu, Jianwei Yang, Jianfeng Gao, Chunyuan Li, and Yong~Jae Lee.
\newblock Gligen: Open-set grounded text-to-image generation.
\newblock {\em arXiv preprint arXiv:2301.07093}, 2023.

\bibitem{lin14microsoft}
Tsung{-}Yi Lin, Michael Maire, Serge~J. Belongie, James Hays, Pietro Perona, Deva Ramanan, Piotr Doll{\'{a}}r, and C.~Lawrence Zitnick.
\newblock Microsoft {COCO:} common objects in context.
\newblock In {\em Proceedings of the European Conference on Computer Vision ({ECCV})}, 2014.

\bibitem{lin2014microsoft}
Tsung-Yi Lin, Michael Maire, Serge Belongie, James Hays, Pietro Perona, Deva Ramanan, Piotr Doll{\'a}r, and C~Lawrence Zitnick.
\newblock Microsoft coco: Common objects in context.
\newblock In {\em Proceedings of the European Conference on Computer Vision ({ECCV})}, 2014.

\bibitem{liu2022compositional}
Nan Liu, Shuang Li, Yilun Du, Antonio Torralba, and Joshua~B Tenenbaum.
\newblock Compositional visual generation with composable diffusion models.
\newblock In {\em Computer Vision--ECCV 2022: 17th European Conference, Tel Aviv, Israel, October 23--27, 2022, Proceedings, Part XVII}, pages 423--439. Springer, 2022.

\bibitem{OWL-ViT}
Matthias Minderer, Alexey Gritsenko, Austin Stone, Maxim Neumann, Dirk Weissenborn, Alexey Dosovitskiy, Aravindh Mahendran, Anurag Arnab, Mostafa Dehghani, Zhuoran Shen, et~al.
\newblock Simple open-vocabulary object detection with vision transformers.
\newblock {\em arXiv preprint arXiv:2205.06230}, 2022.

\bibitem{nichol2021glide}
Alex Nichol, Prafulla Dhariwal, Aditya Ramesh, Pranav Shyam, Pamela Mishkin, Bob McGrew, Ilya Sutskever, and Mark Chen.
\newblock Glide: Towards photorealistic image generation and editing with text-guided diffusion models.
\newblock {\em arXiv preprint arXiv:2112.10741}, 2021.

\bibitem{park2019semantic}
Taesung Park, Ming-Yu Liu, Ting-Chun Wang, and Jun-Yan Zhu.
\newblock Semantic image synthesis with spatially-adaptive normalization.
\newblock In {\em Proceedings of the IEEE/CVF conference on computer vision and pattern recognition}, pages 2337--2346, 2019.

\bibitem{Flickr30K-entities}
Bryan~A. Plummer, Liwei Wang, Christopher~M. Cervantes, Juan~C. Caicedo, Julia Hockenmaier, and Svetlana Lazebnik.
\newblock Flickr30k entities: Collecting region-to-phrase correspondences for richer image-to-sentence models.
\newblock {\em International Journal of Computer Vision ({IJCV})}, 2017.

\bibitem{radford2021learning}
Alec Radford, Jong~Wook Kim, Chris Hallacy, Aditya Ramesh, Gabriel Goh, Sandhini Agarwal, Girish Sastry, Amanda Askell, Pamela Mishkin, Jack Clark, et~al.
\newblock Learning transferable visual models from natural language supervision.
\newblock In {\em International conference on machine learning}, pages 8748--8763. PMLR, 2021.

\bibitem{Dalle-v2}
Aditya Ramesh, Prafulla Dhariwal, Alex Nichol, Casey Chu, and Mark Chen.
\newblock Hierarchical text-conditional image generation with clip latents.
\newblock {\em arXiv preprint arXiv:2204.06125}, page~3, 2022.

\bibitem{ramesh2021zero}
Aditya Ramesh, Mikhail Pavlov, Gabriel Goh, Scott Gray, Chelsea Voss, Alec Radford, Mark Chen, and Ilya Sutskever.
\newblock Zero-shot text-to-image generation.
\newblock In {\em International Conference on Machine Learning}, pages 8821--8831. PMLR, 2021.

\bibitem{reed2016generative}
Scott Reed, Zeynep Akata, Xinchen Yan, Lajanugen Logeswaran, Bernt Schiele, and Honglak Lee.
\newblock Generative adversarial text to image synthesis.
\newblock In {\em International conference on machine learning}, pages 1060--1069. PMLR, 2016.

\bibitem{stable_diffusion}
Robin Rombach, Andreas Blattmann, Dominik Lorenz, Patrick Esser, and Bj{\"o}rn Ommer.
\newblock High-resolution image synthesis with latent diffusion models.
\newblock In {\em Proceedings of the {IEEE} Conference on Computer Vision and Pattern Recognition ({CVPR})}, pages 10684--10695, 2022.

\bibitem{rombach2022high}
Robin Rombach, Andreas Blattmann, Dominik Lorenz, Patrick Esser, and Bj{\"o}rn Ommer.
\newblock High-resolution image synthesis with latent diffusion models.
\newblock In {\em Proceedings of the IEEE/CVF conference on computer vision and pattern recognition}, pages 10684--10695, 2022.

\bibitem{unet}
Olaf Ronneberger, Philipp Fischer, and Thomas Brox.
\newblock U-net: Convolutional networks for biomedical image segmentation.
\newblock In {\em MICCAI}, 2015.

\bibitem{dreambooth}
Nataniel Ruiz, Yuanzhen Li, Varun Jampani, Yael Pritch, Michael Rubinstein, and Kfir Aberman.
\newblock Dreambooth: Fine tuning text-to-image diffusion models for subject-driven generation.
\newblock {\em arXiv preprint arXiv:2208.12242}, 2022.

\bibitem{saharia2022photorealistic}
Chitwan Saharia, William Chan, Saurabh Saxena, Lala Li, Jay Whang, Emily Denton, Seyed Kamyar~Seyed Ghasemipour, Raphael Gontijo-Lopes, Burcu~Karagol Ayan, Tim Salimans, Jonathan Ho, David~J. Fleet, and Mohammad Norouzi.
\newblock Photorealistic text-to-image diffusion models with deep language understanding.
\newblock In Alice~H. Oh, Alekh Agarwal, Danielle Belgrave, and Kyunghyun Cho, editors, {\em Advances in Neural Information Processing Systems}, 2022.

\bibitem{Laion-5b}
Christoph Schuhmann, Romain Beaumont, Richard Vencu, Cade Gordon, Ross Wightman, Mehdi Cherti, Theo Coombes, Aarush Katta, Clayton Mullis, Mitchell Wortsman, et~al.
\newblock Laion-5b: An open large-scale dataset for training next generation image-text models.
\newblock {\em arXiv preprint arXiv:2210.08402}, 2022.

\bibitem{singh2023high}
Jaskirat Singh, Stephen Gould, and Liang Zheng.
\newblock High-fidelity guided image synthesis with latent diffusion models.
\newblock In {\em Proceedings of the IEEE/CVF Conference on Computer Vision and Pattern Recognition}, pages 5997--6006, 2023.

\bibitem{sun2019image}
Wei Sun and Tianfu Wu.
\newblock Image synthesis from reconfigurable layout and style.
\newblock In {\em Proceedings of the IEEE/CVF International Conference on Computer Vision}, pages 10531--10540, 2019.

\bibitem{sylvain2021object}
Tristan Sylvain, Pengchuan Zhang, Yoshua Bengio, R~Devon Hjelm, and Shikhar Sharma.
\newblock Object-centric image generation from layouts.
\newblock In {\em Proceedings of the AAAI Conference on Artificial Intelligence}, pages 2647--2655, 2021.

\bibitem{tao2022df}
Ming Tao, Hao Tang, Fei Wu, Xiao-Yuan Jing, Bing-Kun Bao, and Changsheng Xu.
\newblock Df-gan: A simple and effective baseline for text-to-image synthesis.
\newblock In {\em Proceedings of the IEEE/CVF Conference on Computer Vision and Pattern Recognition}, pages 16515--16525, 2022.

\bibitem{valevski2022unitune}
Dani Valevski, Matan Kalman, Yossi Matias, and Yaniv Leviathan.
\newblock Unitune: Text-driven image editing by fine tuning an image generation model on a single image.
\newblock {\em arXiv preprint arXiv:2210.09477}, 2022.

\bibitem{xie2023boxdiff}
Jinheng Xie, Yuexiang Li, Yawen Huang, Haozhe Liu, Wentian Zhang, Yefeng Zheng, and Mike~Zheng Shou.
\newblock Boxdiff: Text-to-image synthesis with training-free box-constrained diffusion.
\newblock {\em arXiv preprint arXiv:2307.10816}, 2023.

\bibitem{xu2018attngan}
Tao Xu, Pengchuan Zhang, Qiuyuan Huang, Han Zhang, Zhe Gan, Xiaolei Huang, and Xiaodong He.
\newblock Attngan: Fine-grained text to image generation with attentional generative adversarial networks.
\newblock In {\em Proceedings of the IEEE conference on computer vision and pattern recognition}, pages 1316--1324, 2018.

\bibitem{yang2022modeling}
Zuopeng Yang, Daqing Liu, Chaoyue Wang, Jie Yang, and Dacheng Tao.
\newblock Modeling image composition for complex scene generation.
\newblock In {\em Proceedings of the IEEE/CVF Conference on Computer Vision and Pattern Recognition}, pages 7764--7773, 2022.

\bibitem{yang2022reco}
Zhengyuan Yang, Jianfeng Wang, Zhe Gan, Linjie Li, Kevin Lin, Chenfei Wu, Nan Duan, Zicheng Liu, Ce Liu, Michael Zeng, et~al.
\newblock Reco: Region-controlled text-to-image generation.
\newblock {\em arXiv preprint arXiv:2211.15518}, 2022.

\bibitem{Flickr30k}
Peter Young, Alice Lai, Micah Hodosh, and Julia Hockenmaier.
\newblock From image descriptions to visual denotations: New similarity metrics for semantic inference over event descriptions.
\newblock {\em TACL}, 2014.

\bibitem{yu2022scaling}
Jiahui Yu, Yuanzhong Xu, Jing~Yu Koh, Thang Luong, Gunjan Baid, Zirui Wang, Vijay Vasudevan, Alexander Ku, Yinfei Yang, Burcu~Karagol Ayan, et~al.
\newblock Scaling autoregressive models for content-rich text-to-image generation.
\newblock {\em arXiv preprint arXiv:2206.10789}, 2022.

\bibitem{zhang2021cross}
Han Zhang, Jing~Yu Koh, Jason Baldridge, Honglak Lee, and Yinfei Yang.
\newblock Cross-modal contrastive learning for text-to-image generation.
\newblock In {\em Proceedings of the IEEE/CVF conference on computer vision and pattern recognition}, pages 833--842, 2021.

\bibitem{zhang2017stackgan}
Han Zhang, Tao Xu, Hongsheng Li, Shaoting Zhang, Xiaogang Wang, Xiaolei Huang, and Dimitris~N Metaxas.
\newblock Stackgan: Text to photo-realistic image synthesis with stacked generative adversarial networks.
\newblock In {\em Proceedings of the IEEE international conference on computer vision}, pages 5907--5915, 2017.

\bibitem{zhang2018stackgan++}
Han Zhang, Tao Xu, Hongsheng Li, Shaoting Zhang, Xiaogang Wang, Xiaolei Huang, and Dimitris~N Metaxas.
\newblock Stackgan++: Realistic image synthesis with stacked generative adversarial networks.
\newblock {\em IEEE transactions on pattern analysis and machine intelligence}, 41(8):1947--1962, 2018.

\bibitem{zhang2023adding}
Lvmin Zhang and Maneesh Agrawala.
\newblock Adding conditional control to text-to-image diffusion models.
\newblock {\em arXiv preprint arXiv:2302.05543}, 2023.

\bibitem{zhao2019image}
Bo Zhao, Lili Meng, Weidong Yin, and Leonid Sigal.
\newblock Image generation from layout.
\newblock In {\em Proceedings of the IEEE/CVF Conference on Computer Vision and Pattern Recognition}, pages 8584--8593, 2019.

\bibitem{Detic}
Xingyi Zhou, Rohit Girdhar, Armand Joulin, Philipp Kr{\"a}henb{\"u}hl, and Ishan Misra.
\newblock Detecting twenty-thousand classes using image-level supervision.
\newblock In {\em Proceedings of the European Conference on Computer Vision ({ECCV})}, 2022.

\end{thebibliography}
}
\clearpage

\appendix
\renewcommand{\thetable}{A\arabic{table}}
\renewcommand{\thefigure}{A\arabic{figure}}
\setcounter{figure}{0}  
\setcounter{figure}{0}  

This appendix contains the following parts:
\begin{itemize}
    \item \textbf{Implementation Details.} We provide more details of the experimental settings, including the network architecture and noise scheduler.

    \item \textbf{Evaluation Dataset and Metrics.} We provide the details of dataset and evaluation metrics used in the experiments part.

    \item \textbf{Ablation Study.} A detailed quantitative evaluation is presented to understand the impact of various components and hyper-parameter selections. We investigate the influence of guided steps, layer-specific losses, and the loss scale factor for backward guidance.  %

  \item \textbf{Analysis on Initial Noise.} We demonstrate that different prompts with the same initial noise generate images with similar layouts. Therefore, a good choice of initial noise is essential for the success of guidance. Additionally, we quantitatively prove that using the defined loss on cross-attention allows for optimal initial noise selection, enhancing guidance performance.

    \item \textbf{Analysis on Different Tokens.} We visualize the cross-attention map of different prompts and provide extra experiments about controlling the layout of the generated image with only padding tokens.

    \item \textbf{More Examples.} We provide additional examples of our method, including examples under VISOR \cite{visor} protocol and real image editing examples.

\end{itemize}

\section{Implementation Details}
\label{s:Imp_details}

We provide additional details of our experimental settings.
\paragraph{Network Architecture.} In all experiments, we use the Stable Diffusion (SD) V-1.5 \cite{stable_diffusion} as our base model without any architecture modification. The diffusion model is trained in the latent space of an autoencoder. Specifically, the diffusion model adopts the U-Net \cite{unet} architecture with a relative downsampling factor of 8. The down-sampling branch of the U-Net has three sequential cross-attention blocks. The mid part of the U-Net has only one cross-attention block. The up-sampling branch of the U-Net has three sequential cross-attention blocks. In each cross-attention block, there are repeated layers following the order: ResBlock $\rightarrow$ Self-Attention $\rightarrow$ Cross-Attention. The cross-attention blocks in the down-sampling branch, mid part, and up-sampling have 2, 1, and 3 such repeated patterns, respectively. 

\paragraph{Noise Scheduler.} The LMSDscheudler is utilized in all of our experiments with 51 time steps and beta values starting at 0.00085 and ending at 0.012, following a linear scheduler. We also adopt class-free guidance, as suggested in~\cite{ho2022classifier}, with a guidance scale of 7.5, consistent with prior work~\cite{stable_diffusion}.

\section{Evaluation Datasets and Metrics}

\paragraph{VISOR~\cite{visor}.}

We follow the evaluation process described in \cite{visor} to compute the VISOR metric, which is designed to quantify the spatial understanding abilities of text-to-image models.
This metric focuses on two-dimensional relationships, such as \textit{left}, \textit{right}, \textit{above}, and \textit{below}, between two objects. 
We measure object accuracy (OA), which is the probability that the generated image contains both objects specified in the text prompt. $\rm{VISOR}_{uncond}$ is the probability that generating both objects with correct spatial relationship, and $\rm{VISOR}_{cond}$ is the conditional probability of correct spatial relationships being generated, given that both objects were generated correctly.
To generate text prompts for evaluation, we use the 80 object categories from the MS COCO dataset \cite{lin2014microsoft}, resulting in a total of 80 $\times$ 79 $\times$ 4 $=$ 25,280 prompts considering any combination of two object categories for each spatial relationship. 
For each prompt, we generate a single image.
As layout guidance inputs we split the image canvas into two, vertically or horizontally, to create two adjacent bounding boxes depending on the type of spatial relationship defined by the text prompt. 
This only imposes a weak constraint on the layout and can be done automatically (no user intervention is required). 
For a fair comparison to previous methods that are evaluated in \cite{visor}, we use the same detection model (OWL-ViT \cite{OWL-ViT}) as in \cite{visor} when computing the VISOR metric.

\paragraph{COCO 2014~\cite{lin2014microsoft}}
We randomly sampled 1000 images with their annotations for evaluation from the COCO 2014 validation dataset. The bounding boxes in COCO 2014 are not always grounded in the corresponding caption. Therefore, we append the object labels to the caption as the text prompt for image generation following a similar setting in ~\cite{couairon2023zero, avrahami2023spatext}. Besides, we only pick one to three bounding boxes with areas covering at least 5\% of the image panel per sample following the setting in \cite{couairon2023zero}. To assess the quality of the generated images we compute the FID score between the sampled 1000 images from COCO and generated images. We use an open-vocabulary object detector (Detic~\cite{Detic}) to obtain the respective grounding on generated images, which allows quantifying \emph{layout fidelity} using common detection metrics such as average precision (AP). The vocabulary of the detector is constrained to all the COCO object labels.

\paragraph{Flickr30K Entities~\cite{Flickr30K-entities}}
Finally, we evaluate our method on the Flickr30k Entities dataset~\cite{Flickr30K-entities, Flickr30k}, which contains image-caption pairs.
Since the dataset provides visual groundings of the textual descriptions, we sample a single caption per image and its corresponding bounding boxes and use this as input to perform layout-controlled guidance with SD.
We generate a total of 1,000 images using samples from the validation set.
Similarly to the metric used in COCO 2014, we compute the FID score between the original images and the generated ones and use AP as a metric of layout control.
To enhance the reliability of the detector, we convert each phrase in the Flickr30 dataset into a single noun (\eg, \textit{ball}) and filter out unrelated nouns, resulting in a total of 303 categories.
For each image, the target vocabulary for Detic is defined by the grounded entities in the corresponding caption. 
To avoid contaminating the evaluation process with perceived human attributes (such as gender, age, occupation, etc.), we also convert all instances of people (man, woman, child, boy, girl, policeman, student, etc.) to the super-class ``person'' in the target vocabulary for Detic.
Since then the \textit{person} category is predominant, we calculate average precision separately for this category (AP$_\mathrm{p})$ but also report the mean average precision across all categories (mAP).

\section{Ablation Study}
\label{s:ablation_study}

\begin{table} [h]
  \footnotesize
  \newcommand{\xpm}[1]{{\tiny$\pm#1$}}
\centering
\setlength{\tabcolsep}{3.5pt}
\begin{tabular}{ccccc}
  \toprule
  Guidance Step & FID ($\downarrow$) & $\text{AP}_{p}$ ($\uparrow$) &mAP ($\uparrow$) & Inference Time\\
  \midrule
      0 & 76 & 19.4 & 8.7 & $\sim$ 4sec/image\\
      2 & \textbf{81.2} & 29.7 & 13.7 & $\sim$ 4sec/image\\
      5 & 81.4 & 30.3 & 15.6 & $\sim$ 6 sec/image \\
      10 & 82.0 & 33.5 & \textbf{16.7} & $\sim$ 8 sec/image \\ 
      15 & 82.3  & 35.5 & 14.7 & $\sim$ 10 sec/image \\
      20 & 83.2  & 35.6 & 15.3 & $\sim$ 12 sec/image \\
      30 & 83.5  & \textbf{35.7} & 15.3 & $\sim$ 15 sec/image \\
  \bottomrule
\end{tabular}
\caption{Ablation study on guidance steps. }
\label{table:ablation_iteration_step}
\end{table}
\begin{table*} [t!]
  \footnotesize
  \newcommand{\xpm}[1]{{\tiny$\pm#1$}}
\centering
\setlength{\tabcolsep}{3.5pt}
\begin{tabular}{cccccccc@{\hskip 0.5cm}ccc}
  \toprule
  Base Model &Down-1 & Down-2 & Down-3 & Mid-1 & Up-1 & Up-2 & Up-3 & FID ($\downarrow$) & AP$\mathrm{_{p}}$ ($\uparrow$) &mAP ($\uparrow$) \\
  \midrule
       \multirow{11}{*}{\rotatebox[origin=c]{90}{Stable Diffusion \cite{stable_diffusion}}} & 
       \checkmark & \checkmark & \checkmark & ~ & ~ & ~ & ~ & 81.3 & 31.1 & 13.2 \\
        & \checkmark & ~ & ~ & ~ & ~ & ~ & ~ & 83.5 & 23.1 & 10.0 \\
        &  ~ & \checkmark & ~ & ~ & ~ & ~ & ~ & 82.0 & 24.0 & 10.9 \\
        &  ~ & ~ & \checkmark & ~ & ~ & ~ & ~ & 82.2 & 34.5 & 14.2 \\
        &  ~ & ~ & ~ & \checkmark & ~ & ~ & ~ & 82.1 & 30.0 & 15.2 \\
        &  ~ & ~ & ~ & \checkmark & \checkmark & ~ & ~ & 82.0 & 33.5 & \textbf{16.7} \\
        &  ~ & ~ & ~ & \checkmark & ~ & \checkmark & ~ & 86.3 & 30.9 & 14.0 \\
        &  ~ & ~ & ~ & \checkmark & ~ & ~ & \checkmark & 84.1 & 23.5 & 10.5 \\
        &  ~ & ~ & ~ & ~ & \checkmark & \checkmark & \checkmark & 84.5 & 35.6 & 16.5 \\
        &  ~ & ~ & ~ & ~ & \checkmark & ~ & ~ & \textbf{81.2} & \textbf{36.0} & 15.1 \\
        &  ~ & ~ & ~ & ~ & ~  & \checkmark & ~ & 87.5 & 35.0  & 14.3 \\
        &  ~ & ~ & ~ & ~ & ~ & ~ & \checkmark & 85.0 & 25.6 & 9.8 \\
  \bottomrule
\end{tabular}
\caption{Ablation study of loss constraints on different layers.}
\label{table:supp_layer}
\end{table*}
\begin{table} [t]
  \footnotesize
  \newcommand{\xpm}[1]{{\tiny$\pm#1$}}
\centering
\setlength{\tabcolsep}{3.5pt}
\begin{tabular}{cccc}
  \toprule
  Loss Scale & FID ($\downarrow$) & AP$\mathrm{_{p}}$ ($\uparrow$) &mAP ($\uparrow$) \\
  \midrule
      5 & 82.5 & 28.3 & 12.4 \\
        10  & 82.0 & 30.0  & 14.5 \\
        20 & \textbf{81.1} & 34.7 & 15.4 \\
      30  & 82.0 & 33.5 & \textbf{16.7} \\
            50  & 83.8 & \textbf{35.8} & 15.6 \\
            100  & 88.4 & 34.9  & 14.3 \\
            200  & 99.2 & 32.2 & 13.8 \\
            500  & 129.7 & 26.2 & 9.2 \\

  \bottomrule
\end{tabular}
\caption{Ablation study of the loss scale factor.}
\label{table:supp_loss_scale}
\end{table}

In this section, we supplement the ablation studies in the main paper with quantitative evaluations, studying the impact of the guided steps, loss scale factor, and the effect of backward guidance on different layers of the denoising network.
We followed the same setting as described above and in \Cref{s:exp_setup} (main paper) using 1000 captions and their corresponding bounding boxes from the Flickr30K Entities \cite{Flickr30K-entities} dataset to generate images with a pre-specified layout.

\paragraph{Impact of Guidance Step.}

Firstly, we explore the effects of guided steps we perform in the diffusion process. The results are shown in \cref{table:ablation_iteration_step}, we evaluate image quality (FID), AP$_\mathrm{p}$,  layout control (mAP) while varying the number of \textit{guided} steps.
We found no improvement in mAP after 10 steps, and FID gradually deteriorates. We hypothesize that this decline may result from potentially shifting the latent vector away from the distribution that corresponds to the original text embedding. Besides, we could see that when increasing the guided steps in the diffusion process, the computation time increases. This is a trade-off question. Generally, a range of 2-10 guidance steps suffices, but users can fine-tune this based on their specific requirements.

\paragraph{Impact of Layers.}
Secondly, we study the behavior of different layers, by applying backward guidance on the cross-attention maps across different layers of the network. 
The results are shown in \Cref{table:supp_layer}. As stated in \Cref{s:analysis} and illustrated in the table, layers of the down-sampling branch are the least likely to conform to layout control (with Down-1 $<$ Down-2 $<$ Down-3 in terms of mAP). %
In general, high-resolution blocks (such as Down-1 or Up-3) should not be used to control the layout.
To achieve the best trade-off between image quality and layout control, a combination of the mid-block (Mid-1) and the first cross-attention block in up-sampling branch (Up-1) of the U-Net is the optimal choice overall.

\paragraph{Impact of Loss Scale Factor.}
We follow the same setup to evaluate the scale factor $\eta$ used as the strength of the loss for backward guidance. In \Cref{table:supp_loss_scale} we report the FID, AP$_\mathrm{p}$ and mAP for different loss scale factors. When the loss scale is set to 5--50, the FID is low compared to a larger loss scale factor, indicating that the quality with a loss scale factor of 5--50 is generally good. To achieve better control over the layout, the loss scale factors of 20--50 have the lowest AP$_\mathrm{p}$ and mAP. According to the experiments, a loss scale factor of 20--50 works generally well.
This factor can be adjusted by the user to get more realistic images or achieve better control over the layout.

\section{Analysis on Initial Noise}

\begin{figure}[t]
    \centering
    \includegraphics[width=\columnwidth]{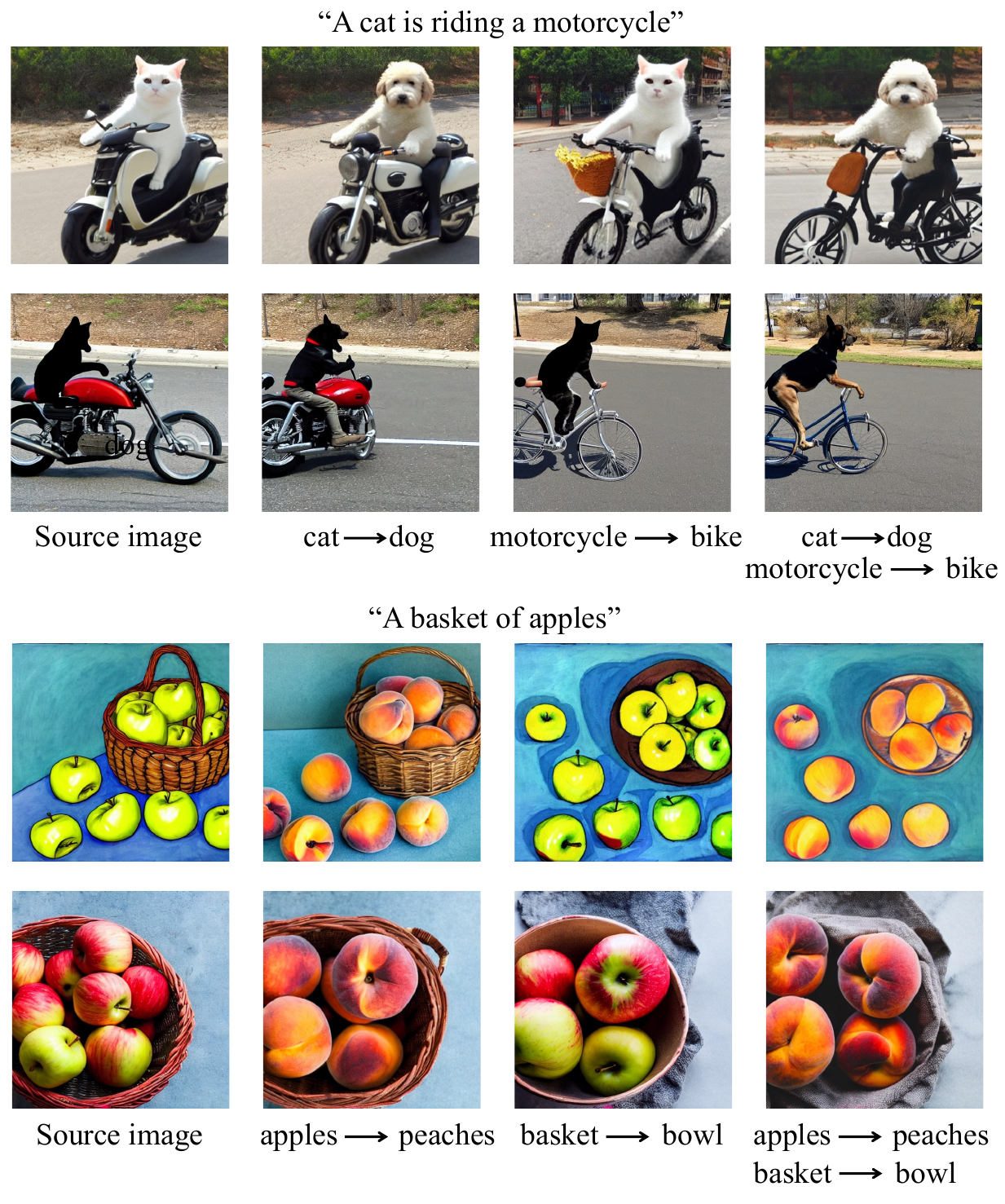}
    \caption{Each row has the same initial noise. We could see that even if we changed the object word in one sentence, the overall layout remains similar.}
    \label{fig:initial_noise}
\end{figure}

\begin{figure}[t]
    \centering
    \includegraphics[width=\columnwidth]{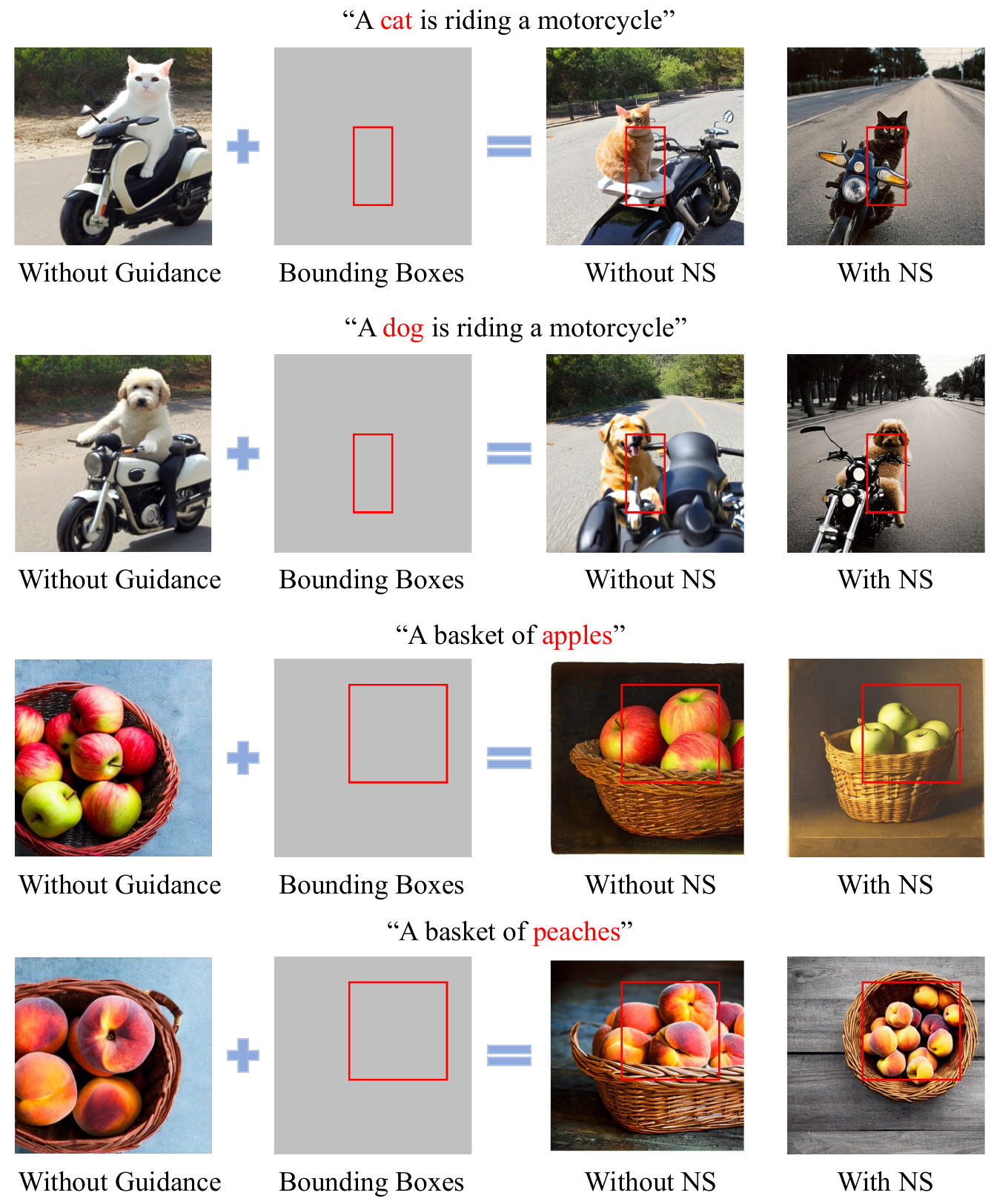}
    \caption{We qualitatively compare the generated results with and without noise selection (NS). The results show that with noise selection, our backward guidance achieves better layout control.}
    \label{fig:noise_selection}
\end{figure}

We conduct an in-depth analysis of the effects of initial noise. As illustrated in ~\Cref{fig:initial_noise}, the initial noise reveals significant spatial information about the layout. Notably, altering sentence words does not affect this final layout significantly. ~\Cref{fig:noise_selection} offers a visual comparison of scenarios with and without noise selection. The results indicate that our backward guidance achieves better control when noise selection is employed. Furthermore, ~\Cref{table:noise_selection} quantitatively assesses the impact of noise selection on COCO 2014 and Flickr30K datasets. Methods incorporating noise selection consistently outperform others, underscoring the efficacy of our loss as a noise selection metric.

\section{Analysis on Different Tokens}
\label{s:analysis_on_token}

Next, we study the type of information carried by different tokens and their corresponding cross-attention maps, which is relevant for layout guidance.

\begin{figure*}[t]
    \centering
    \includegraphics[width=0.96\linewidth]{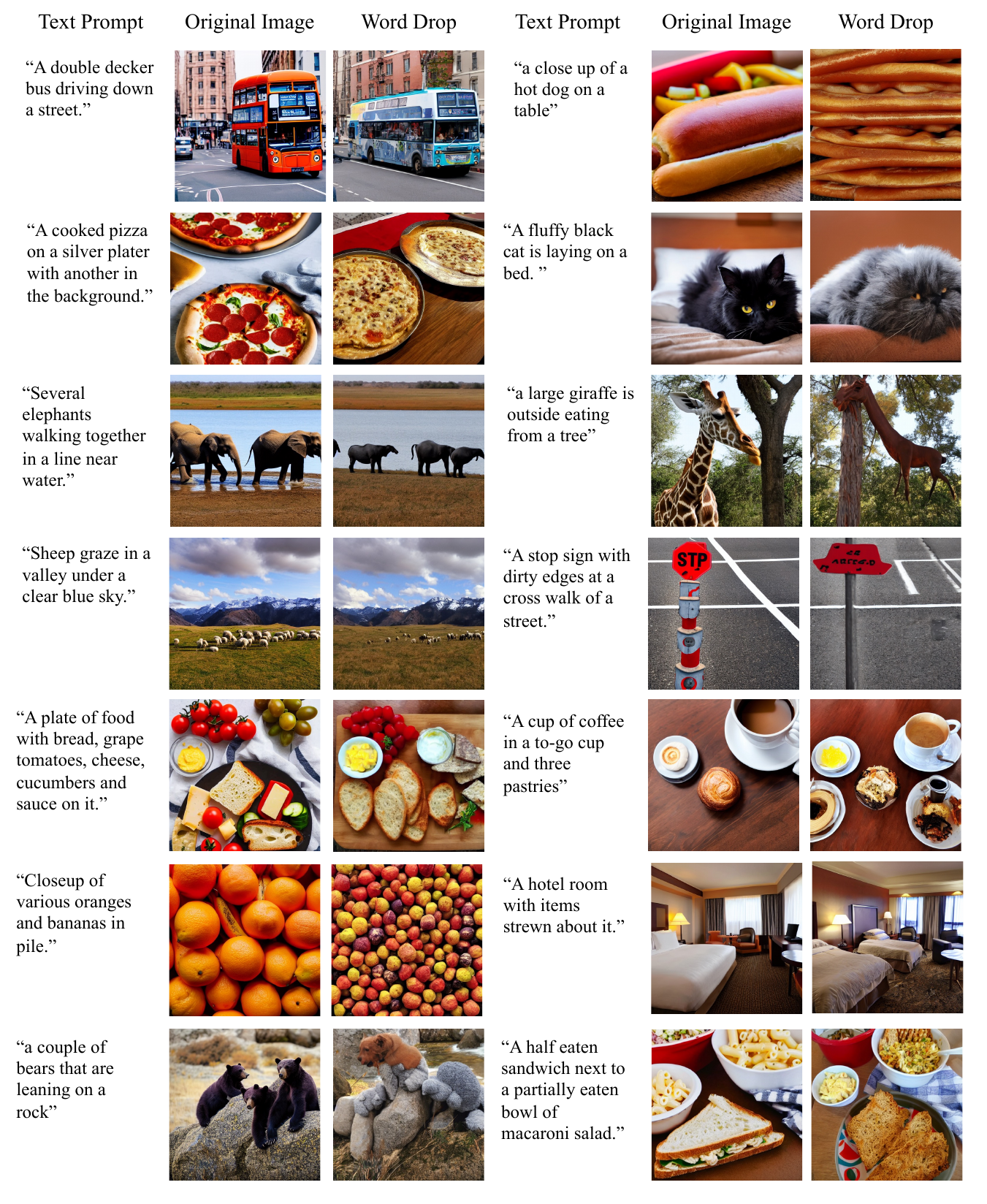}
    \caption{Generating images without ``seeing'' the full-text prompt. We replace the token embeddings for all words in each caption with their \textit{padding} token embeddings (word drop). We observe that the generated images after word dropping exhibit similar semantics and layout to the images generated from the full-text prompt, suggesting that significant information about the image is contained in padding tokens.}
    \label{fig:word_drop}
\end{figure*}

\paragraph{Removing Word Tokens.} We first show that the \textit{padding} tokens convey a significant amount of semantic information. In \Cref{fig:word_drop}, we randomly pick a subset of captions from MSCOCO \cite{lin2014microsoft} and generate images using the Stable Diffusion model and the full caption as the input prompt. 
As a comparison, after the captions pass through the text encoder, we replace the token embeddings of each caption with the embeddings of its corresponding padding tokens, thus creating a prompt that consists only of padding tokens. Then, we use this prompt to generate images. 
Surprisingly, despite only generating from padding (\ie, non-word) token embeddings, we observe that the generated images (Word Drop in \Cref{fig:word_drop}) closely follow both the semantics and the layout of the image generated from the full-text prompt.
Thus, the figure clearly demonstrates that the padding tokens contain the information of the whole sentence. 
This further justifies why in forward guidance padding tokens cannot be ignored, \ie, it would be insufficient to attempt to control selected word tokens only (main paper, \Cref{fig:forward_vs_backward}). 
In backward guidance, however, controlling the cross-attention maps of padding tokens is not necessary; this is now done by back-propagating and updating the latent, which subsequently changes the cross-attention maps of all tokens, even those that are not explicitly controlled. 

\begin{figure*}[t]
    \centering
    \includegraphics[width=\linewidth]{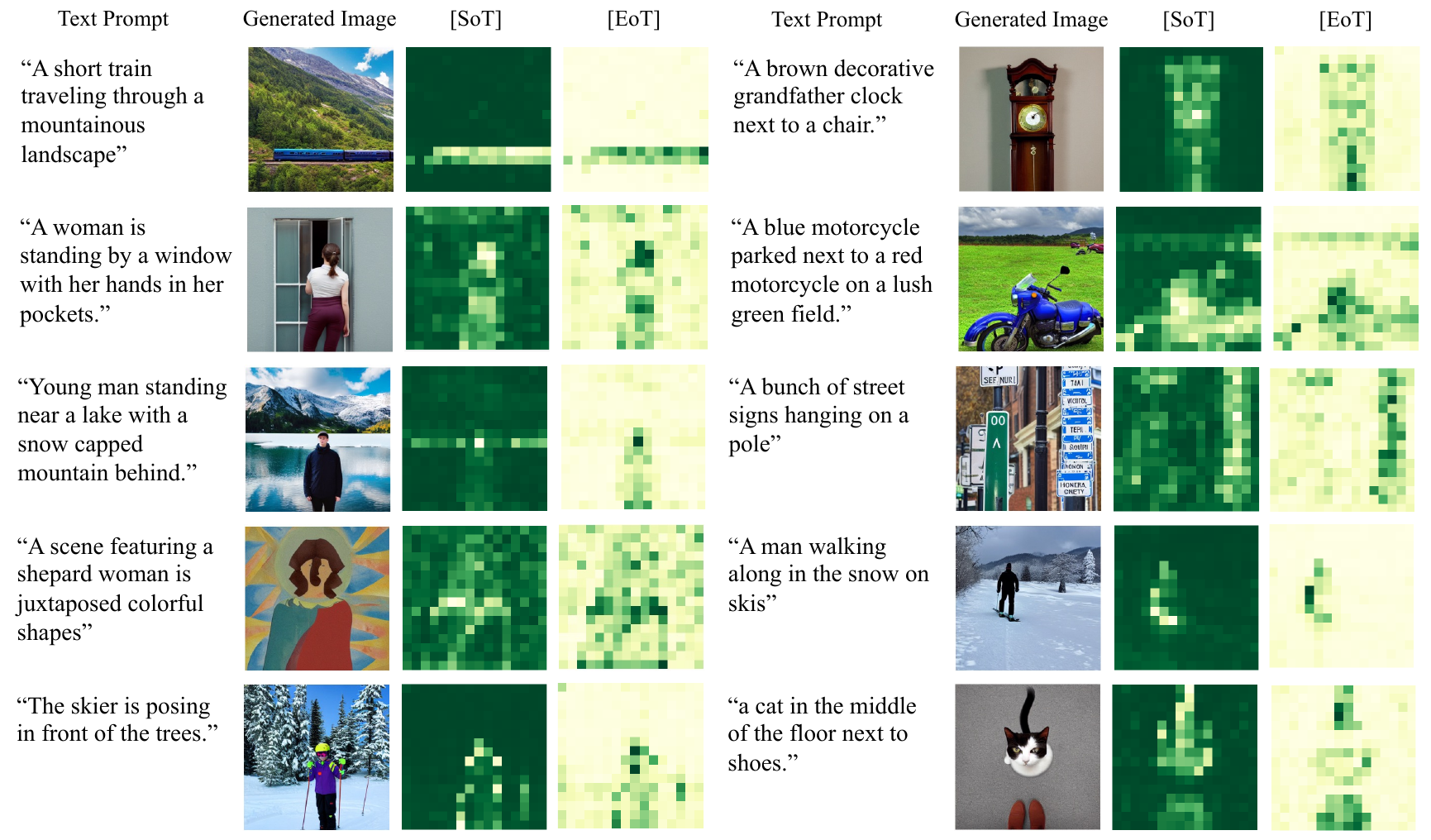}
    \caption{Visualization of cross-attention maps of start token (\texttt{[SoT]}) and padding tokens (\texttt{[EoT]}) at the final step of inference. Cross-attention maps are taken from the first cross-attention block of the up-sampling branch of U-Net and averaged over all attention heads.}
    \label{fig:padding_ca_visual}
\end{figure*}

\paragraph{Cross-Attention Maps of Special Tokens.} 
During our experiments, we found that the cross-attention of the padding tokens has a strong connection to the foreground of the generated images. We illustrated this in \Cref{fig:padding_token} (main paper), which shows that the cross-attention maps of padding tokens resemble saliency maps, while the cross-attention maps of the start tokens are mostly complementary to those of padding tokens (\ie, they capture what can be considered as background). 
In \Cref{fig:padding_ca_visual}, we show more examples of the cross-attention maps of the \textit{start} and \textit{padding} tokens. The captions are randomly taken from MSCOCO \cite{lin14microsoft}. 
This figure further highlights the observation that cross-attention maps of these special tokens contain important semantic and spatial information. For example, in the first row,  given ``A short train traveling through a mountainous landscape" as the input prompt, the cross-attention map of the padding tokens aligns with the generated train and the start token focuses on the background of the generated image.

\begin{figure*}[t]
    \centering
    \includegraphics[width=\linewidth]{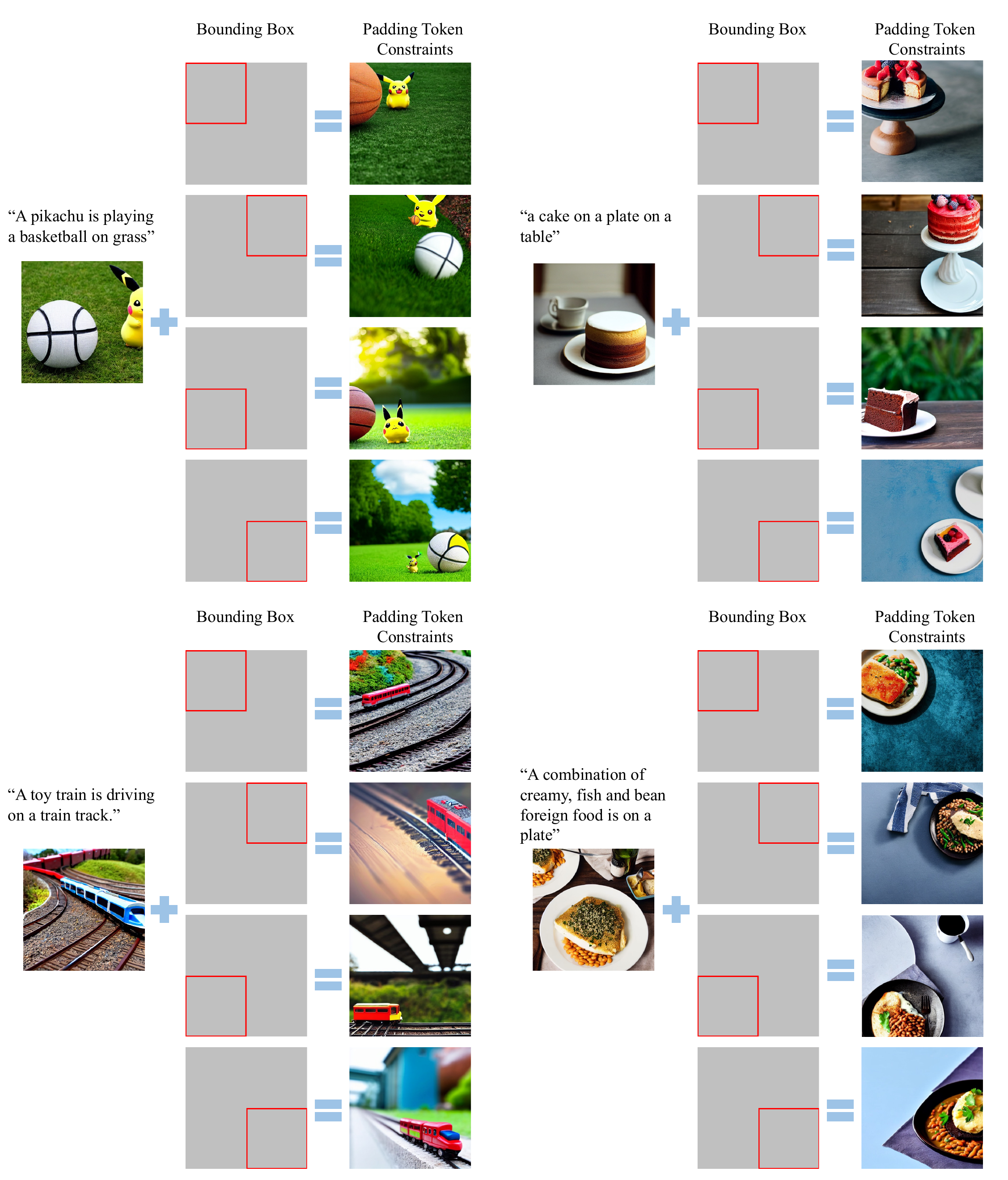}
    \caption{Backward guidance \emph{only} on the \emph{padding} tokens. We observe that the cross-attention of padding tokens typically represents the foreground of the generated image. Therefore, by spatially guiding the cross-attention maps that correspond to padding tokens, we can control the position of the foreground, which may include multiple objects (e.g., ``pikachu'' and ``basketball'').}
    \label{fig:only_padding_token}
\end{figure*}

\paragraph{Layout Control with Only Padding Tokens.}  
Motivated by the examples above, we perform backward guidance only on the cross-attention maps of padding tokens to control the spatial layout of all foreground objects simultaneously (as a group). Some examples are shown in \Cref{fig:only_padding_token}. 
This figure verifies our assumption that by guiding the cross-attention map of the padding tokens alone one can control the composition of the images at the foreground/background level. %

\begin{table} [t!]
  \footnotesize
  \newcommand{\xpm}[1]{{\tiny$\pm#1$}}
\centering
\setlength{\tabcolsep}{3.5pt}
\scalebox{0.95}{\begin{tabular}{lcccccc}
  \toprule
  \multirow{2}{*}{Base Model}  & \multirow{2}{*}{NS} & \multicolumn{2}{c}{COCO 2014} & \multicolumn{3}{c}{Flickr30K} \\
           \cmidrule{3-7}
  & & FID ($\downarrow$) &mAP ($\uparrow$)  & FID ($\downarrow$) &mAP ($\uparrow$) & AP$_\mathrm{P}$ ($\uparrow$)  \\
    
  \midrule

    Stable Diffusion & \XSolidBrush  & 74.4 & 33.6 & 82.0 & 33.5 & 16.7 \\
    Stable Diffusion & \checkmark  & \textbf{73.3} & \textbf{35.7} & \textbf{78.9 }& \textbf{35.6 }& \textbf{17.9 }\\

  \bottomrule
\end{tabular}}
\caption{Ablation Study on Noise Selection (NS).}
\label{table:noise_selection}
\end{table}

\section{More Examples.}
\label{s:more_examples}

\begin{figure*}[t]
    \centering
    \includegraphics[width=\linewidth]{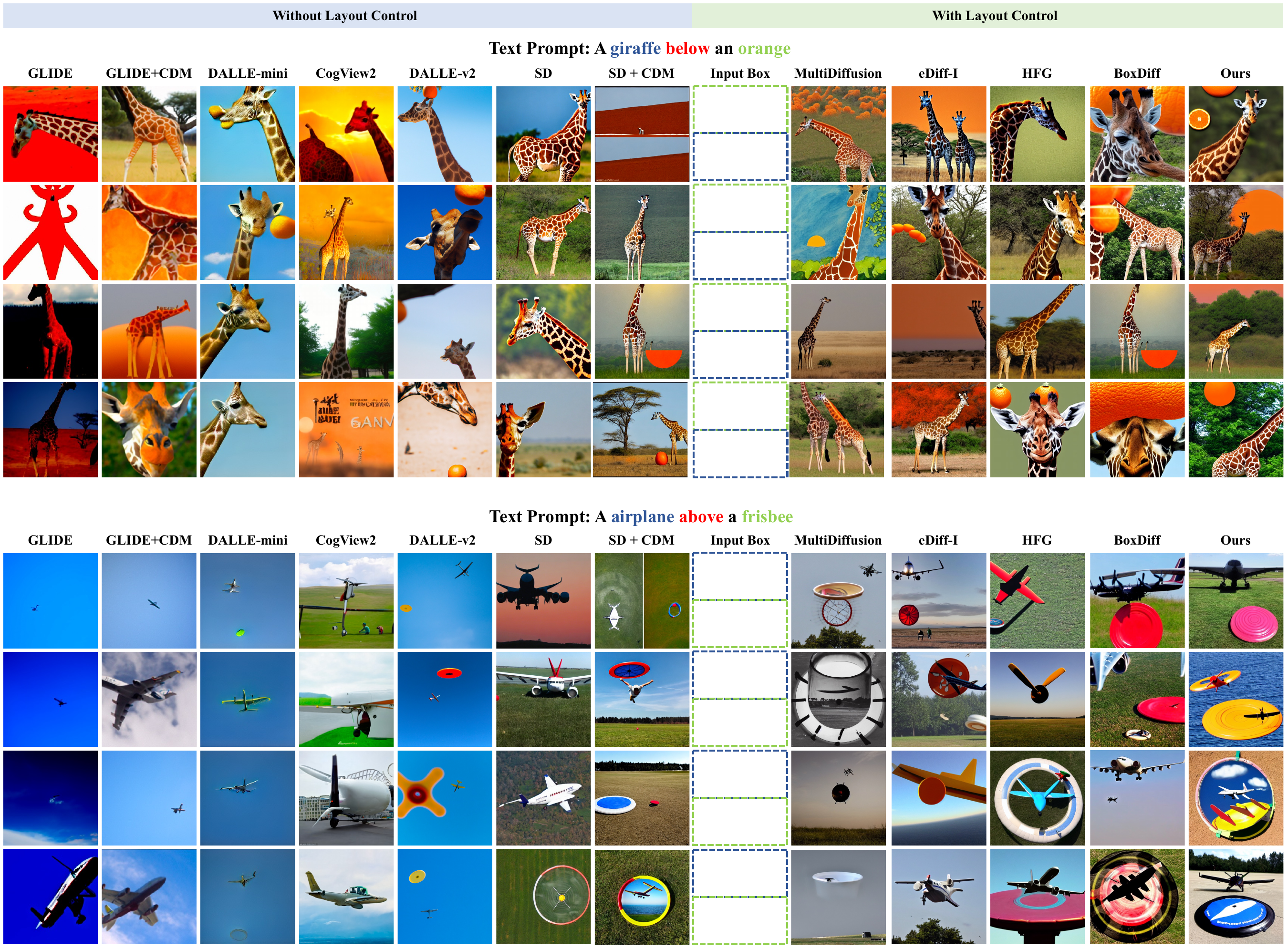}
    \caption{Qualitative comparison between different generative models. For each prompt, we generate four images. Some images of other models are from the demo website of \cite{visor}.}
    \label{fig:additional_visor}
\end{figure*}

\begin{figure*}[t]
    \centering
    \includegraphics[width=\linewidth]{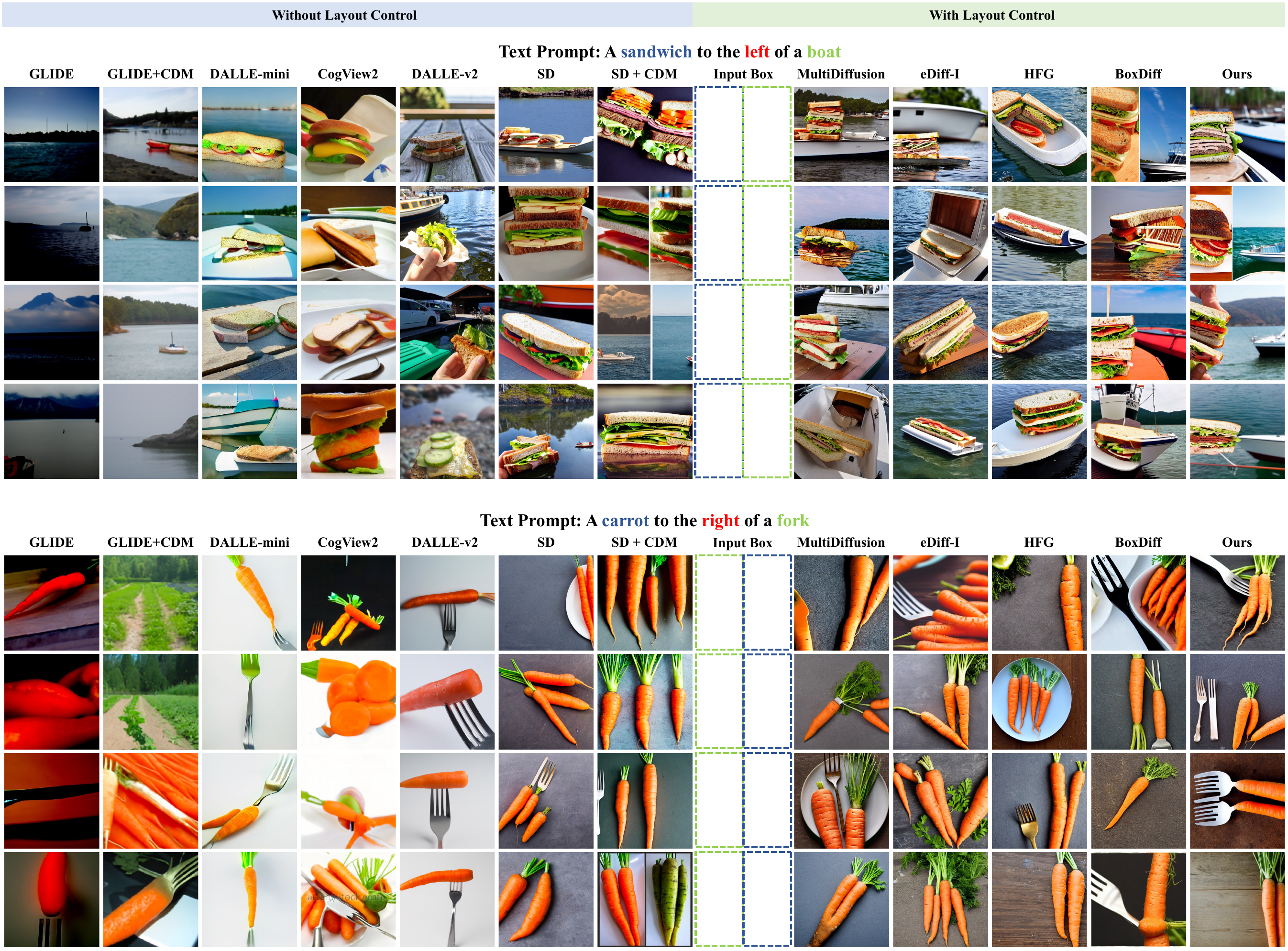}
    \caption{Qualitative comparison between different generative models. For each prompt, we generate four images. Some images of other models are from the demo website of \cite{visor}.}
    \label{fig:additional_visor_2}
\end{figure*}

\paragraph{More Examples under VISOR Protocol.}
We show more examples under the VISOR protocol in \Cref{fig:additional_visor} and \Cref{fig:additional_visor_2}. Our method generates the correct spatial relationships as shown in the figures. There are also some failure cases, such as the last row in \Cref{fig:additional_visor}. Our method fails to generate both a fork and a carrot. This is an inherited problem from the Stable Diffusion model. 
However, in most cases, layout guidance helps generate \emph{all} entities in the text prompt, even when the unguided Stable Diffusion fails (\eg, as is often the case with atypical scene compositions), as well as conforming to a specific spatial arrangement.    

\begin{figure*}[t]
    \centering
    \includegraphics[width=0.95\linewidth]{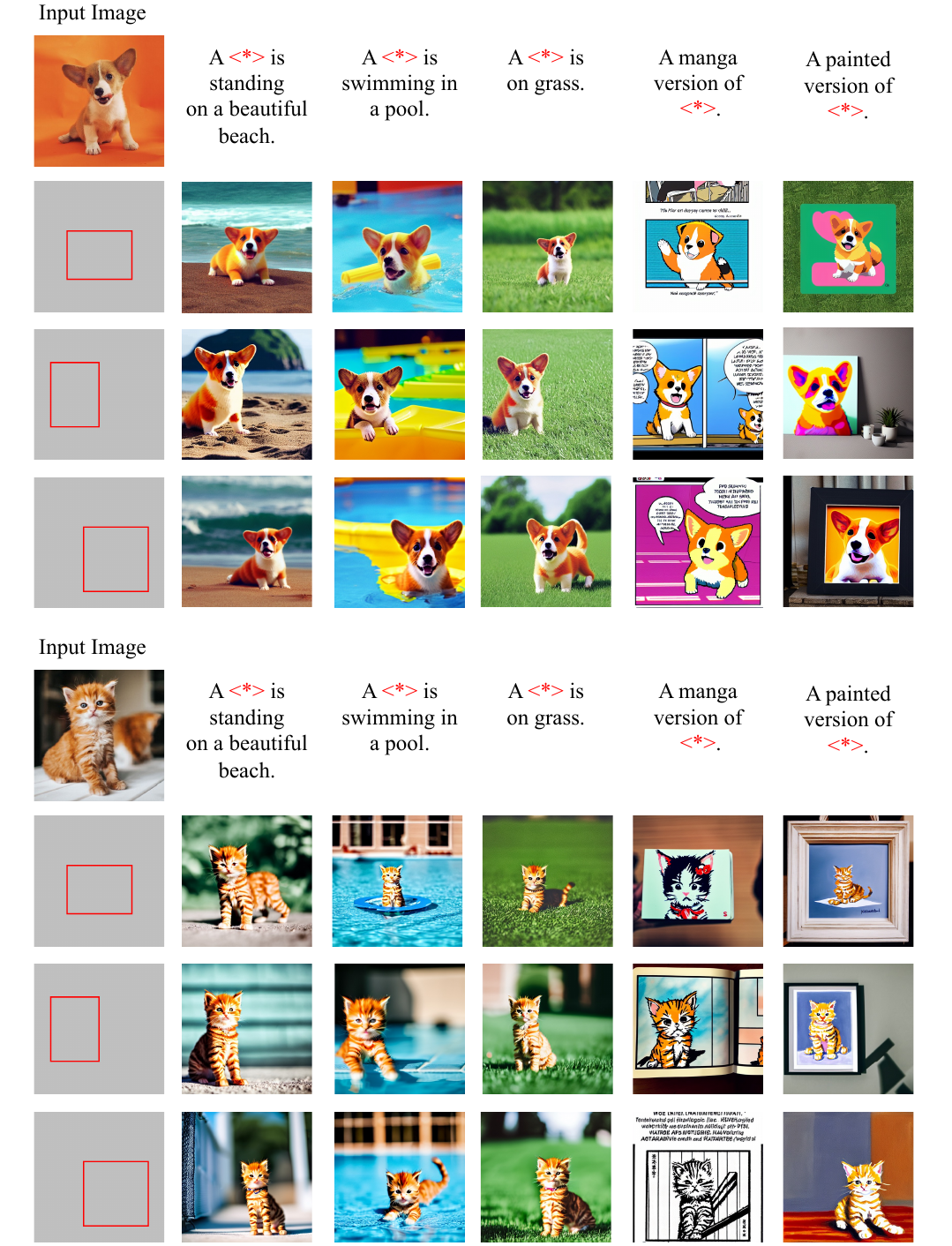}
    \caption{More examples of real image editing. \textcolor{red}{$\langle*\rangle$ } is the learned token that encodes the object in the real image.}
    \label{fig:supp_real_image}
\end{figure*}

\paragraph{More Image Editing Examples.}
We show more examples of real image editing in \Cref{fig:supp_real_image}. Specifically, we train for 500 steps to learn the embedding of \textcolor{red}{$\langle*\rangle$} with text inversion and then 150 steps fine-tuning of the text encoder and denoiser network with Dreambooth. After finalizing the model, we perform inference with our backward guidance using different text prompts and user-specified bounding boxes. As shown in the figure, we manage to change the context, layout, and style of the given real image.

\end{document}